\documentclass{article}

\usepackage{graphicx}
\usepackage{subcaption}
\usepackage[normalem]{ulem}
\graphicspath{ {./images/} }

\usepackage[nonatbib,preprint]{neurips_2022}

\usepackage[utf8]{inputenc} % allow utf-8 input
\usepackage[T1]{fontenc}    % use 8-bit T1 fonts
\usepackage{hyperref}       % hyperlinks
\usepackage{url}            % simple URL typesetting
\usepackage{booktabs}       % professional-quality tables
\usepackage{amsfonts}       % blackboard math symbols
\usepackage{nicefrac}       % compact symbols for 1/2, etc.
\usepackage{microtype}      % microtypography
\usepackage{xcolor}         % colors
\usepackage[flushleft]{threeparttable} %table notes
\title{C$^3$Fusion: Consistent Contrastive Colon Fusion, 
Towards Deep SLAM in Colonoscopy}

\author{%
  Erez Posner \\
  Intuitive Surgical\\
  \texttt{Erez.Posner@intusurg.com} \\
    \And
    Adi Zholkover \\
    Intuitive Surgical\\
    \texttt{Adi.Zholkover@intusurg.com} \\
    \And
    Netanel Frank \\
    Intuitive Surgical\\
    \texttt{Netanel.Frank@intusurg.com} \\
    \And
    Moshe Bouhnik \\
    Intuitive Surgical\\
    \texttt{Moshe.Bouhnik@intusurg.com} \\
}

\begin{document}

\maketitle

\begin{abstract}
3D colon reconstruction from Optical Colonoscopy (OC) to detect non-examined surfaces remains an unsolved problem. The challenges arise from the nature of optical colonoscopy data, characterized by highly reflective low-texture surfaces, drastic illumination changes and frequent tracking loss. 
Recent methods demonstrate compelling results, but suffer from: (1) frangible frame-to-frame (or frame-to-model) pose estimation resulting in many tracking failures; or (2) rely on point-based representations at the cost of scan quality. In this paper, we propose a novel reconstruction framework that addresses these issues end to end, which result in both quantitatively and qualitatively accurate and robust 3D colon reconstruction. Our SLAM approach, which employs correspondences based on contrastive deep features, and deep consistent depth maps, estimates globally optimized poses, is able to recover from frequent tracking failures, and estimates a global consistent 3D model; all within a single framework. We perform an extensive experimental evaluation on multiple synthetic and real colonoscopy videos, showing high-quality results and comparisons against relevant baselines.

\end{abstract}

\section{Introduction}
The third most commonly diagnosed cancer worldwide is colorectal cancer (CRC) with over than 1.9 million incident cases in 2020~\cite{ccfs}. CRC is also among the most preventable cancers~\cite{ijcm113278} and can be prevented from progressing if detected in it's primary stages by conducting screening and early detection measures~\cite{zjrms93934,Holakouie-Naini}. Consequently, global incidence rates have been decreasing in the screening-eligible age group (50–75) due to the adoption of CRC
screening~\cite{10.3389/fonc.2021.730689}. The most common screening procedure is optical colonoscopy (OC)~\cite{Liang2010}, which visually inspects the mucosal surface for abnormalities in the colon such as colorectal lesions. Nevertheless, performing a thorough endoscopic colon investigation solely from OC is very difficult. In practice, this means that not all regions of the colon will be covered and fully examined; consequently, tainting the polyp detection rate. Lately, we are seeing a bloom in deep learning-based methods adapted to predict depth-maps from OC~\cite{zhang2021colde,AppearanceFlow,GANDepth,SLAMEndoscopyGAN}, aimed at providing a complete 3D geometric information of the colon including polyps. Thus, indicating the un-inspected surfaces during OC; as a result, increasing the polyp detection rate. 

Despite the profusion of reconstruction solutions, a holistic solution for the problem of 3D colon reconstruction at scale that addresses real life issues during OC has yet to be seen. This is due to the numerous requirements that such system would have to support:

\textit{Accurate depth prediction} - producing high-quality geometry-consistent depth estimation from a monocular video is imperative as well as challenging. \textit{Scalability} - chosen representation should support extended scale environments while preserving global structure, and high local accuracy. \textit{Global consistency} - the method should be robust to pose drifts and estimation error in order to enable the re-examination of previously scanned areas or loop closure. \textit{Robust camera tracking} - tracking failure is extremely frequent in OC. Occlusions, fast motions, featureless regions~\cite{RNNSLAM} and deficient frames are a fraction of the reasons that contribute to loss of track. When these occur, the system should have the ability to re-localize the camera position.

There have been studies addressing specific parts of these problems~\cite{AppearanceFlow,OZYORUK2021102058,RNNSLAM,DBLP:journals/corr/abs-2107-13263,YAO2021102180,zhang2021lighting}. Direct SLAM systems optimize a photometric error which is susceptible to drastic illumination changes in OC imagery. Ma et al.~\cite{RNNSLAM} reconstructed fragments of the colon using Direct Sparse Odometry (DSO)~\cite{DSO} and a Recurrent Neural Network (RNN) for depth estimation. Zhang et al.~\cite{zhang2021lighting} predicted gamma correction value to alleviate sudden illumination changes and~\cite{zhang2021colde} improved the depth estimation network using geometry-consistency losses. Indirect SLAM methods, like~\cite{DBLP:journals/corr/DaiNZIT16,DBLP:journals/corr/Mur-ArtalMT15}, usually utilize keypoints matching based on handcrafted descriptors. This kind of descriptors e.g., SIFT~\cite{Lowe:2004uq}, are based on local gradients and hence not well suited to often texture-less and shadow prone OC imagery. Modern deep network based descriptors adopt CNN to predict both keypoint and descriptors for local feature matching. DeTone et al.~\cite{super_point}, predicts keypoints and descriptors directly from a pre-trained DNN. However, it's localization accuracy is hampered due to the low dimensionality output. Moreover, as it's training is based on corner detection keypoints it is not optimized to OC cases, characterized with numerous occlusions. Although these aforementioned studies show promising results, there hasn't been a single solution to tackle all of these requirements up to date.

Our goal in this paper is to rigorously cope with \textit{all} of these requirements in a single, end-to-end 3D reconstruction pipeline. At the core of our method is a robust positioning estimation scheme that utilizes contrastive deep-feature based correspondences. The proposed method globally optimizes the camera pose per-frame, taking into consideration all previously captured frames in an effective \textit{local-to-global} hierarchical optimization framework.

In summary, the main contributions of our work are as follows: (1) A novel, deep-learning-driven global pose alignment SLAM system for OC which incorporates the complete sequence of input frames and removes the fuzzy nature of temporal tracking accuracy issues; (2)  Large-scale colon 3D reconstruction, demonstrating model refinement in revisited areas, recovery from tracking failures, and robustness to drift and continuous loop closures; and (3) a novel method for local feature matching in low-texture areas, implicit loop closures in highly indistinguishable environments and highly-accurate fine-scale pose alignment.

\begin{figure}[htbp]
\centering
\includegraphics[width=1\textwidth]{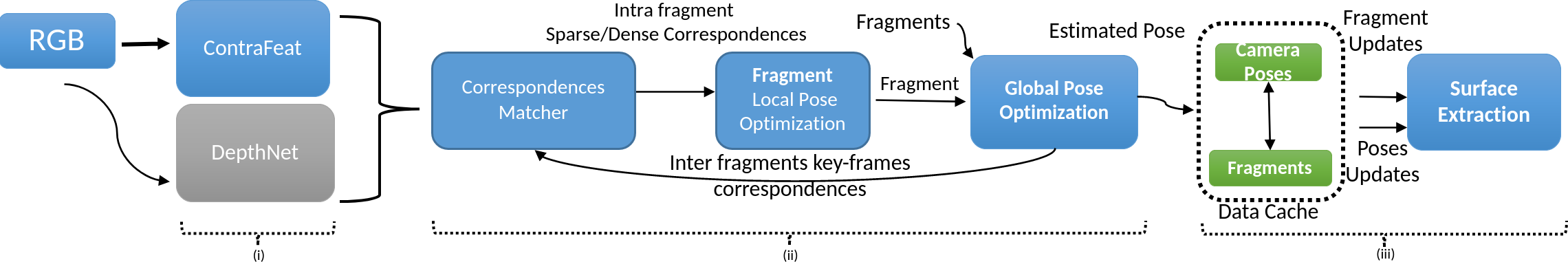}
\caption{Our novel, deep-learning-driven global pose alignment framework for colonoscopy SLAM system} \label{fig1}
\end{figure}

\section{Method overview}
\label{method_overview}
The main system pipeline (shown in Fig.~\ref{fig1}) consists of three major parts: ($i$) depth estimation and deep feature extraction, ($ii$) hierarchical pose optimisation, and ($iii$) surface fusion. 
For each new frame, part ($i$) outputs a depth-map and keypoints with their deep descriptors, by inferencing DepthNet (Sec.~\ref{deepdepth}) and ContraFeat (Sec.~\ref{Deep_Descriptors}) respectively. Part ($ii$) starts with matching the new keypoints against previous frames and filtering mismatches (Sec.~\ref{feature_corr}) to improve alignment and avoid false loop closures. To manage large scaled sequences comprised out of massive amount of frames and to make pose alignment fast, we carry out a hierarchical local-to-global pose optimization. This achieves robustness to frequent loss of tracking as we do not solely depend on temporal consistency. Thereby, enabling swift re-localization and allowing multiple visits of the same regions within the scene.
% To achieve robustness to frequent loss of tracking, instead of solely depending on temporal consistency, We apply hierarchical pose optimization (Sec.~\ref{pose_optim}). Thus, enabling swiftly re-localization and allowing multiple visits of the same regions within the scene.

% To manage large scaled sequences comprised out of massive amount of frames, and to make pose alignment fast, we carry out a hierarchical local-to-global pose optimization.
On the first (local) hierarchy level, $fragments$ are composed of sets of successive frames sharing similar spatial coverage. Each frame's pose is optimized by taking all of the $fragment's$ frames into account. On the second (global) hierarchy level, all $fragments'$ pose are optimized with respect to each other. In part ($iii$) the global 3D scene representation is acquired by fusing all fragments (Sec.~\ref{3d_reco}) into a non-parametric surface represented implicitly by a scalable truncated signed distance function (TSDF)~\cite{conf/siggraph/CurlessL96,6751168} following with marching cubes~\cite{10.1145/37402.37422}  applied to this volume to extract the final mesh. 
% (i.e., model continuous surfaces, systematically
% regularize noise and efficiently incremental updates)

% In ($iii$) the global 3D scene representation is also accomplished in a hierarchical two-stage strategy. On the first hierarchy level, for each fragment we obtain dense scene reconstruction using its composed frames. On the second hierarchy level, all fragments are aligned according to their estimated global poses. Thus, if loop closure occurs the fragments position will be updated according to their current global pose. Finally, all fragments are fused (Sec.~\ref{3d_reco}) into a non-parametric surface represented implicitly by a scalable truncated signed distance function (TSDF)~\cite{conf/siggraph/CurlessL96,6751168} following with marching cubes~\cite{10.1145/37402.37422}  applied to this volume to extract the final mesh.

\section{Deep-Depth \& Deep-Descriptors}
\subsection{Deep-Depth self-supervised training}\label{deepdepth}
Given as input an RGB image $I_t$, DepthNet predicts a depth map $\tilde{D_t}$. During training, for every frame $I_t$ in a sequence of three sequential frames {$I_{t-1}, I_t, I_{t+1}$}, DepthNet predicts their corresponding depth maps {$\tilde{D}_{t-1}, \tilde{D}_t, \tilde{D}_{t+1}$}. PoseNet predicts relative camera poses between each image pair:  
$T_{ij} \forall{(i,j)\in F_i} = \{\forall i|j=i\pm1\}$. 

The network architecture is similar to the one used in Monodepth2 \cite{monodepth2}. To reduce the impact of strong visual distortions (e.g. lens distortion) that characterized OC videos, we adopt deformable convolution~\cite{Dai2017DeformableCN} for the depth encoder, similar to~\cite{Chen2021DistortionAwareMD, Kumar2020FisheyeDistanceNetSS}. We train in the same manner as \cite{monodepth2}, using auto-masking ($\mu$) and per-pixel minimum photometric loss ($L_{ph}$). Since the photometric loss is not sufficiently informative for low-texture regions, common in OC imagery, and to enforce structural coherence, we apply extra regularization in the form of depth consistency loss~\cite{Bian2019UnsupervisedSD}, and extra spatio-temporal consistency losses. Considering that the additional regularization implicitly imposes depth smoothness, we discard the smoothness term ($L_{ds}$) used in~\cite{monodepth2}.

To deal with specular reflections and occlusions by haustral folds we: (1) mask and in-paint specular reflections as in \cite{OZYORUK2021102058} and (2) remove outlier pixels having a loss greater than the 80-th percentile for the photometric and depth consistency errors.
We compute the additional spatio-temporal consistency and depth consistency losses i.e., $L_{ph-extra}$ and $L_{dc}$ respectively, between all image pairs in $S$. 
\begin{equation}
    L^{(i,j)}_{ph-extra} = \frac{1}{|V_\mu|}\sum_{p\in{V_\mu}}pe(I_i, I_{j \to i}),
\end{equation}
\begin{equation}
    L^{(i,j)}_{dc}(\tilde{D}_{j \to i}, \hat{D}_{i}) = \frac{\left|{\tilde{D}_{j \to i} - \hat{D}_{i}}\right|}{\tilde{D}_{j \to i} + \hat{D}_{i}},
\end{equation}
\[\forall{(i,j)}\in S = \{i=\{t-1,t+1\}, j=\{t-1, t, t+1\}, i\ne{j}\}\]

where $pe$ is the photometric error from \cite{monodepth2} containing L1 and SSIM losses, $V_\mu$ is a set comprises of all valid pixels in mask $\mu$ and the specular reflection mask based on image intensity threshold. $I_{j \to i}$ is a warped view of $I_j$ to $I_i$ pose, using the predicted depth $\tilde{D}_j$ and $T_{ij}$. $\tilde{D}_{j \to i}$ is the predicted depth map for image $I_j$ in the coordinate system of image $I_i$. $\hat{D}_{i}$ is the interpolated depth map from the estimated depth map $\tilde{D}_{i}$. 
% Note that these losses do not handle occlusions as auto-masking for $L_{ph}$ does, since $L_{ph-extra}$ and $L_{dc}$ only use a single warped reference image.
The final loss for the network is given in Eq. \ref{eq:3}, where $\lambda_{ph-extra}$, $\lambda_{dc}$ are weights for the different loss components. 
\begin{equation} \label{eq:3}
L = L_{ph} +\lambda_{ph-extra} \frac{1}{|S|} \sum_{(i, j)\in S}L^{(i,j)}_{ph-extra} + \lambda_{dc} \frac{1}{|S|} \sum_{(i, j)\in S}L^{(i,j)}_{dc} 
\end{equation}

\begin{figure}[hptb]
    \centering
    \begin{tabular}[t]{ccccc}

       \begin{tabular}{c}% if you add [t], than sub images are pushed down
             \parbox[t]{1mm}{\rotatebox[origin=c]{90}{Input}} \\\\\\\\\\\
             \parbox[t]{1mm}{\rotatebox[origin=c]{90}{GT Depth}}  \\\\\\\\
             \parbox[t]{1mm}{\rotatebox[origin=c]{90}{Monodepth2~\cite{monodepth2}}}  \\\\\\\\
             \parbox[t]{1mm}{\rotatebox[origin=c]{90}{Ours}}
        \end{tabular}
        \begin{subfigure}{.2\textwidth}
                \begin{tabular}{c}% if you add [t], than sub images are pushed down
                    \begin{subfigure}[t]{1\textwidth}
                        \centering
                       \includegraphics[width=1\textwidth]{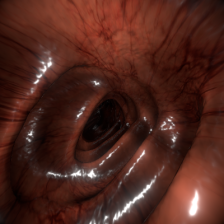}
                    \end{subfigure}\\
                    \begin{subfigure}[t]{1\textwidth}
                        \centering
                        \includegraphics[width=1\textwidth]{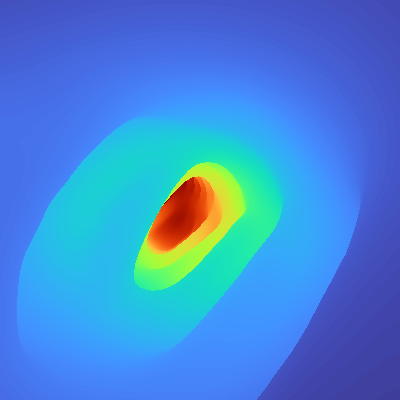}
                    \end{subfigure}\\
                    \begin{subfigure}[t]{1\textwidth}
                        \centering
                        \includegraphics[width=1\textwidth]{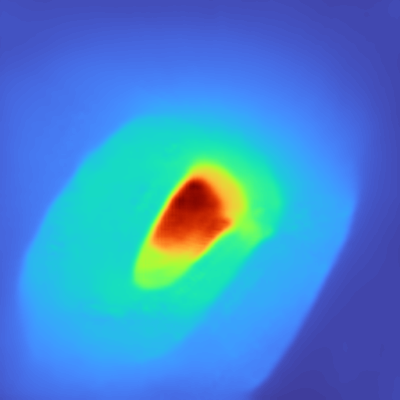}
                    \end{subfigure}\\
                    \begin{subfigure}[t]{1\textwidth}
                        \centering
                        \includegraphics[width=1\textwidth]{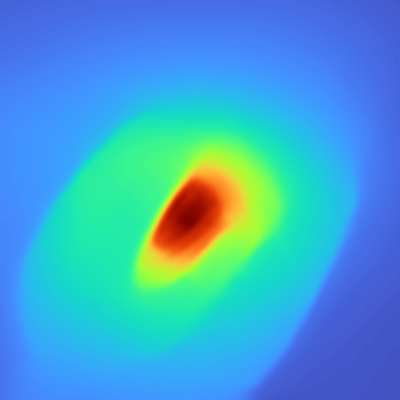}
                    \end{subfigure}
                    
            \end{tabular}
        \end{subfigure}

        \begin{subfigure}{.2\textwidth}
                \begin{tabular}{c}% if you add [t], than sub images are pushed down
                    \begin{subfigure}[t]{1\textwidth}
                        \centering
                        \includegraphics[width=1\textwidth]{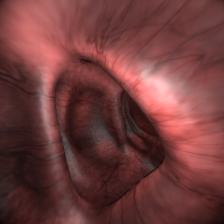}
                    \end{subfigure}\\
                    \begin{subfigure}[t]{1\textwidth}
                        \centering
                        \includegraphics[width=1\textwidth]{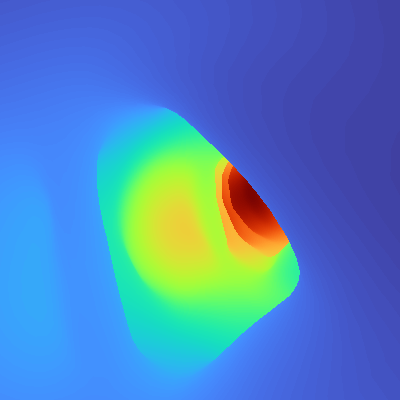}
                    \end{subfigure}\\
                    \begin{subfigure}[t]{1\textwidth}
                        \centering
                        \includegraphics[width=1\textwidth]{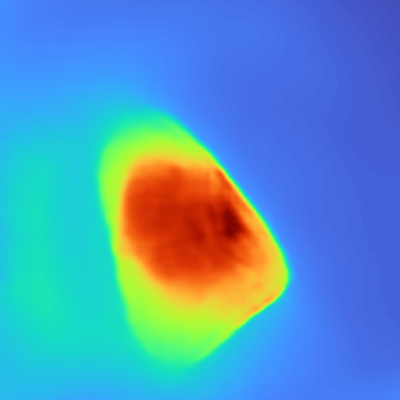}
                    \end{subfigure}\\
                    \begin{subfigure}[t]{1\textwidth}
                        \centering
                        \includegraphics[width=1\textwidth]{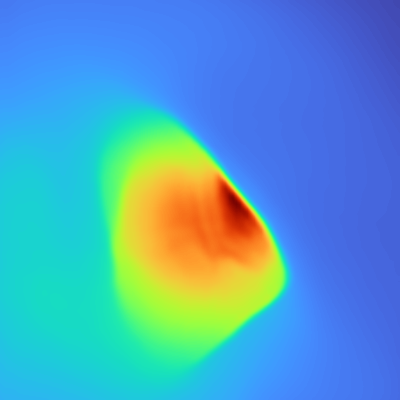}
                    \end{subfigure}
                    
            \end{tabular}
        \end{subfigure}
        \begin{subfigure}{.2\textwidth}
                \begin{tabular}{c}% if you add [t], than sub images are pushed down
                    \begin{subfigure}[t]{1\textwidth}
                        \centering
                        \includegraphics[width=1\textwidth]{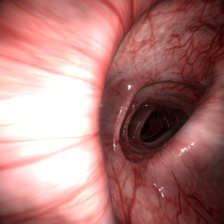}
                    \end{subfigure}\\
                    \begin{subfigure}[t]{1\textwidth}
                        \centering
                        \includegraphics[width=1\textwidth]{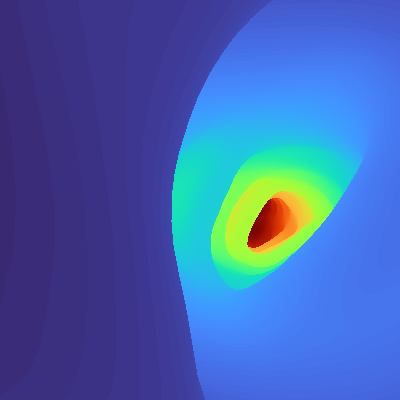}
                    \end{subfigure}\\
                    \begin{subfigure}[t]{1\textwidth}
                        \centering
                        \includegraphics[width=1\textwidth]{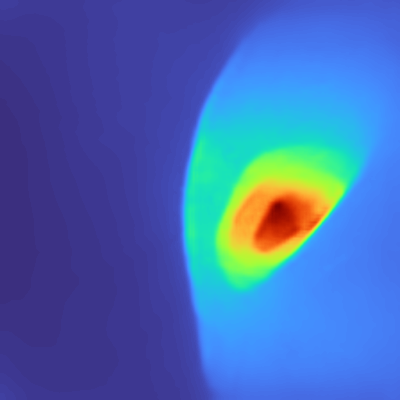}
                    \end{subfigure}\\
                    \begin{subfigure}[t]{1\textwidth}
                        \centering
                       \includegraphics[width=1\textwidth]{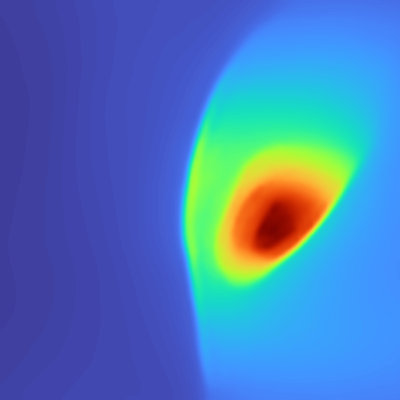}
                    \end{subfigure}
            \end{tabular}
        \end{subfigure}
        \begin{subfigure}{.2\textwidth}
                \begin{tabular}{c}% if you add [t], than sub images are pushed down
                    \begin{subfigure}[t]{1\textwidth}
                        \centering
                        \includegraphics[width=1\textwidth]{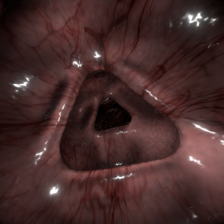}
                    \end{subfigure}\\
                    \begin{subfigure}[t]{1\textwidth}
                        \centering
                        \includegraphics[width=1\textwidth]{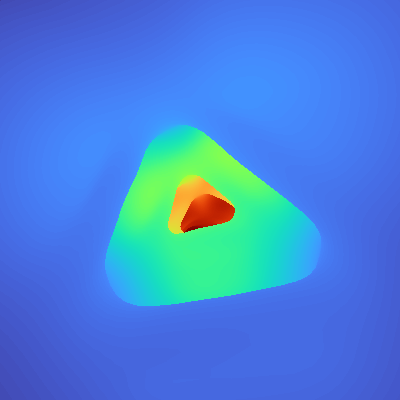}
                    \end{subfigure}\\
                    \begin{subfigure}[t]{1\textwidth}
                        \centering
                        \includegraphics[width=1\textwidth]{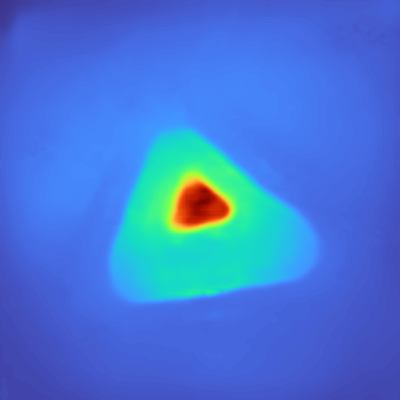}
                    \end{subfigure}\\
                    \begin{subfigure}[t]{1\textwidth}
                        \centering
                        \includegraphics[width=1\textwidth]{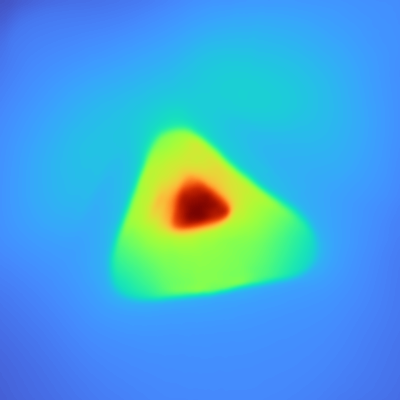}
                    \end{subfigure}
            \end{tabular}
        \end{subfigure}
    \end{tabular}
    \caption{Qualitative comparison of predicted depth-maps on synthetic data.}
\label{fig:monocular_depth_prediction_qualitative}
\end{figure}

\subsection{Deep-Descriptors}\label{Deep_Descriptors}
Our deep feature descriptor block, $ContraFeat$, employs the detected keypoints from~\cite{Lowe:2004uq} in each frame, and extracts their deep feature representations $z = \phi(kp)$, where $kp$ is a SIFT keypoint in 2D pixel coordinates.
For the feature map $\phi$, we use FPN \cite{Lin2017FeaturePN} architecture in the bottom-up stream. The final feature map has the same spatial resolution as the original image and thus, retaining keypoint's pixel level accuracy. Each pixel is represented by a descriptor vector of length $c=128$. Finally, to find correspondence sets between frames, descriptors are matched using cosine similarity. 
To train $ContraFeat$ network, we use synthetic data~\cite{UTS}(see Sec~\ref{results}). Accordingly, keypoints are extracted using~\cite{Lowe:2004uq}. Then, using known depth, pose and camera intrinsics, we collect ground-truth matches from corresponding 3D points and remove occluded points by filtering-out distant matches.
\paragraph{Contrastive loss}Inspired by recent self-supervised learning methods based on contrastive losses~\cite{Sohn2016NPL,Wu2018UnsupervisedFL}, 
%\cite{Chen2020ImprovedBW,Chen2020ASF,Sohn2016NPL,Wu2018UnsupervisedFL}
we use a loss similar to the InfoNCE loss \cite{Oord2018RepresentationLW} to train ContraFeat to learn discriminative representations of keypoints. The contrastive loss for an image pair $(i, j)$ and a correct match $k$ out of $M_{i,j}$ possible matches is given by
\begin{equation}
    l^{i,j,k}_c = -\log\frac{\exp\big((z^k_i)^T \cdot z^k_{j}/\tau\big)} {\sum_{m=0}^{M_{i,j}}\exp\big((z^m_i)^T \cdot z^m_{j}/\tau\big)}
\end{equation}
where $z^m_i$ and $z^m_j$ are descriptor vectors sampled at pixel coordinates $kp^m_i$ and $kp^m_{j}$ in the feature map of images $i$ and $j$, respectively. $\tau$ represents a temperature parameter. We enforce $\Vert{}z^m_i\Vert{}_2$ = 1 via a L2-normalization layer. This loss is then averaged over all ground-truth matches.
\section{Pose Alignment}\label{Pose_Optimization}
Our system takes an RGB-D stream consisting of pairs $(I^i_{RGB},I^i_{D_{pred}})$, where $i$ is the frame index, $I_{RGB}$ is the 3-channel color image and $I_{D_{pred}}$ is the predicted depth map by the DepthNet network. Intrinsic and distortion parameters are assumed to be known. The goal of this stage is to estimate the ideal set of rigid camera poses ${T_i}=\{(R_i,t_i)|R_i \in SO(3),t_i \in \mathbb{R}^3\}_{i=1}^N$ in which all $N$ frames align as best as possible, based on extracted 3D correspondences between all overlapping frames. The estimated transformations $T_{i}(p)= R_ip+t_i$ localizes all frames in the global coordinates system defined relative to the first frame and $p \in \mathbb{R}^3$.  

\subsection{Feature matching}
\label{feature_corr}
%For every new frame $i$, like~\cite{DBLP:journals/corr/TuranAAKS17} we first mask and in-paint specular reflections. Then, SIFT keypoints~\cite{Lowe:2004uq} are detected and their predicted representation is obtained by $E_j = \phi_i(k_j) $, where $\phi_i$ is the features representation by $ContraFeat$. 
%To avoid any false matches between frames, all candidate matches are filtered to produce a frame pairwise correspondences list that will be used in the global pose optimization. 
In order to find correspondence sets that result with a coherent and stable rigid transform between pair of frames $f_i, f_j$, we set to minimize outliers. To this end, we utilize the key point correspondence filter and the Surface Area Filter as in~\cite{DBLP:journals/corr/DaiNZIT16} to filter the sets of frame-pairwise matches based on geometric and feature-representation constancy. The transformation $T_{ij} \in SE(3)$ is constructed between $f_i, f_j$ if minimal 10 matches are found with a re-projection error under 0.02cm and if it is valid. i.e. condition analysis results with a condition number less than $\epsilon_{cn}$.

\subsection{Hierarchical pose optimization} \label{pose_optim}
A colonoscopy procedure typically takes 30-60 minutes at 30 FPS. To be able to process such massive amount of frames in reasonable time we follow~\cite{DBLP:journals/corr/DaiNZIT16} and split the input sequence into fragments of consecutive frames that share similar coverage and apply two stage hierarchical optimization strategy. On the lower hierarchy level, we perform pose-graph optimization~\cite{5681215} to register all frames within a fragment. On the higher hierarchy level, we register all fragments with respect to each fragment's keyframe.

\paragraph{Fragment construction conditions} We keep track of one active fragment at all times. A new frame will either be appended to the active fragment, or will trigger the creation of a new fragment as it keyframe. There are two conditions which determine whether a new fragment should be constructed. (1) Structural affinity between the last two consecutive frames (i.e., minimal number of correspondences found is less than 100) (2) The new frame and the active fragment keyframe view frustums overlap is less than $85\%$.

\paragraph{Inter vs. Intra fragment registration}
The two hierarchies fragment registration processes are similar. In Intra-fragment (local) registration the pose-graph optimization~(see Sec.~\ref{PoseGraph}) is applied for all fragment’s inhabited frames to align as best as possible with respect to the fragment’s keyframe. Whereas in Inter-fragment (global) registration, we estimate the best global registration for all fragments using solely theirs keyframes. 

%The process starts by performing a search and filter for all valid sparse correspondences between all pairs of frames (or keyframes, depending on hierarchy level), followed by Kabsch algorithm~\cite{Kabsch:a12999,gpa} to estimate the relative transform $T_{ij}$ based on the matched features and their predicted depth value.
%followed by a rigid transform estimation for the relative transform $T_{ij}$ using Kabsch algorithm~\cite{Kabsch:a12999,gpa} using the matched features and their depth value from the predicted depth-map.

Note that for the inter-fragment registration, we do not discard keyframes that have no correspondences with past keyframes. Instead, we keep them as a candidates, as they could share correspondences with future fragments. This enables incorporating the lone fragment later on in the sequence.

% \subsubsection{Intra-Fragment and Global Inter-Fragment Registration}\textcolor{blue}{TODO for me - when I removed intra-fragment there is a gap and I cannot simply talk about inter. maybe try to find a way to discuss both hierarchies here.}
% \textcolor{orange}{<<can be removed>> The process of globally aligning all fragments is similar to the intra-fragment optimization other than the fact that it uses the keyframe of each fragment. It's purpose is to estimate the best global registration for all fragments using theirs global keyframes.} Pose graph optimization~\cite{Choi_2015_CVPR}(Sec. 4.3) is applied over all pairwise keyframes transformations estimated from the features matches. Thus, search and filter for all valid sparse correspondences is performed between all pairs of keyframes. \textcolor{orange}{I rewarded this:} Instead of discarding keyframes that have no correspondences with past keyframes, we keep them as a candidates, like~\cite{DBLP:journals/corr/DaiNZIT16}, as they could share correspondences with future fragments . This enables incorporating the lone fragment \textcolor{orange}{\sout {captured data}} later on in the sequence. 

\subsection{Registration as Pose-Graph optimization}  
\label{PoseGraph}
The goal of the pose-graph optimization is to estimate the ideal set of rigid transforms $\mathbb{T}={\{T_i\}}$ such that all set of input frames $F$ (which depends on the hierarchy level) align as best as possible. The process uses~\cite{Kabsch:a12999,gpa} to estimate the relative rigid transforms $T_{ij}\;  \forall (i,j)\in F$ based on the matched features and their predicted depth value. Given $\{T_{ij}\}$, we construct a pose-graph~\cite{5681215} with vertices ${\{f_i\}}$ and edges $T_{ij}$. As in~\cite{Choi_2015_CVPR}, we set to minimize the inconsistency measure $g$ between poses $T_i, T_j$ and the relative pose  $T_{ij}$, defined as the sum of squared distances between corresponding points in $T_i P_i$ and $T_j P_j$:

\begin{equation} \label{eq:1}
g(T_i, T_j, T_{ij}) = \sum_{(i, j)} \|  T_j^{-1} T_i p_i - T_{ij} p_i\|^2 
\end{equation}

Additional outlier removal filter in the form of edges pruning is applied to further improve the algorithm's robustness against false correspondences.

% \subsection{\textcolor{red}{Broken Graph Support/Efficient Optimization Strategy}}
% \label{efficient_optimization}

% Since our goal is to support extended scale environments with tens of thousands of frames in a sequence, our optimization strategy should be efficient and effective. To achieve this, we implemented an optimization strategy that only incorporates frames that are affected by newly observed frame.
% Whenever a new fragment is constructed and correspondences with previous keyframes is performed the newly updated fragments connectivity graph could result with multiple connected-components. Consequently, the  active sub-graph to be optimized is extracted from the connected-components consists the latest keyframe. Hence, only fragment's connected directly/indirectly to the latest keyframe would get affected by it's newly contributing information. The initial guess for pose-graph should be done in globally consistent manner to support future candidates to be connected easily. This is done by a light-complexity and computationally efficient initialization strategy. In this strategy, we fixate the oldest  frame as our anchor node and traverses the pose-graph only once using~\cite{KORF198597}. During the traversal, we approximates each node’s location based on older positioned neighbors, giving similar weight for ego-motion estimation odometry and detecting loop-closure edges.

\section{Scene reconstruction}
\label{3d_reco}
The colons 3D model is reconstructed by carefully fusing all RGB images, their predicted depth maps and the optimized global poses into an implicit scalable TSDF representation. The TSDF's unique features enable us to alleviate any further inconsistencies in successive depth maps predictions. The fusing scheme is based on the premise that the endoscope is slowly being withdrawn during the procedure; consequently, inspected regions won't be visited again. We fuse fragments when enough time has passed since last inspected ($> \epsilon_{f_{na}}$), and when the current camera position is far enough ($> \epsilon_{cf_{d}}$). This approach scales well to non-fixed scenes common in colonoscopy sequences, as demonstrated in Sec.~\ref{qual_res}.

\section{Results}\label{results}
In this section, we analyze the results of our framework on 3 different data-sets: A colon simulator
data-set, a CT colon rigid print, and real optical colonoscopy videos.

We start by evaluating our results using a realistic synthetic colonoscopy simulator~\cite{UTS}. Like ~\cite{zhang2021colde}, we used the simulator to create 8 sequences of endoscopic procedures. For every frame, the ground-truth depth and pose are known. Each sequence contains on average 2000 frames with a trajectory length of 125cm, at a resolution of 512x512 with a field of view (FOV) of 125 degrees. Different 'Material', 'Light' and 'Wetness' properties were set to best resemble real imagery (See appendix~\ref{A. synthetic_data_generation} for details) . We split the data-set sequences, similarly to~\cite{zhang2021colde},  into training/validation/testing sets, containing 9.5k/0.5k/2.5k frames. The metrics below are reported on the test set.

We also captured an endoscopic sequence of a rigid 3D-print CT based colon, using a calibrated Olympus CF-H185L/I  colonoscope, along with the ground-truth camera trajectory that was captured by an EM tracker. The colon 3D model was fabricated as follows: a CT colon scan from~\cite{colonographydata} was segmented in order to extract the 3D surface of the colon, following with post processing re-modeling operations (cleaning, re-topology, skinning etc.) and texturing. The extracted mesh was then printed using a 3D printer. We intend to elaborate on this data-set creation in a future paper.

Lastly, to qualitatively test our framework on real optical colonoscopy sequences, we use Colon10K data-set~\cite{colon10k}. Colon10K data-set contains 20 sequences cropped from full colonoscopies, each contains on average 500 frames. We split them to training/testing sets at a ratio of 80/20\% such that the entire test sequences are not seen during training.
For training and implementation details, please see appendix~\ref{A. imple_details}

\begin{figure}
    \centering
    \begin{tabular}[t]{ccc}
        \begin{subfigure}{.55\textwidth}
            \centering
            \smallskip
            \includegraphics[width=1\linewidth,height=.5\textwidth]{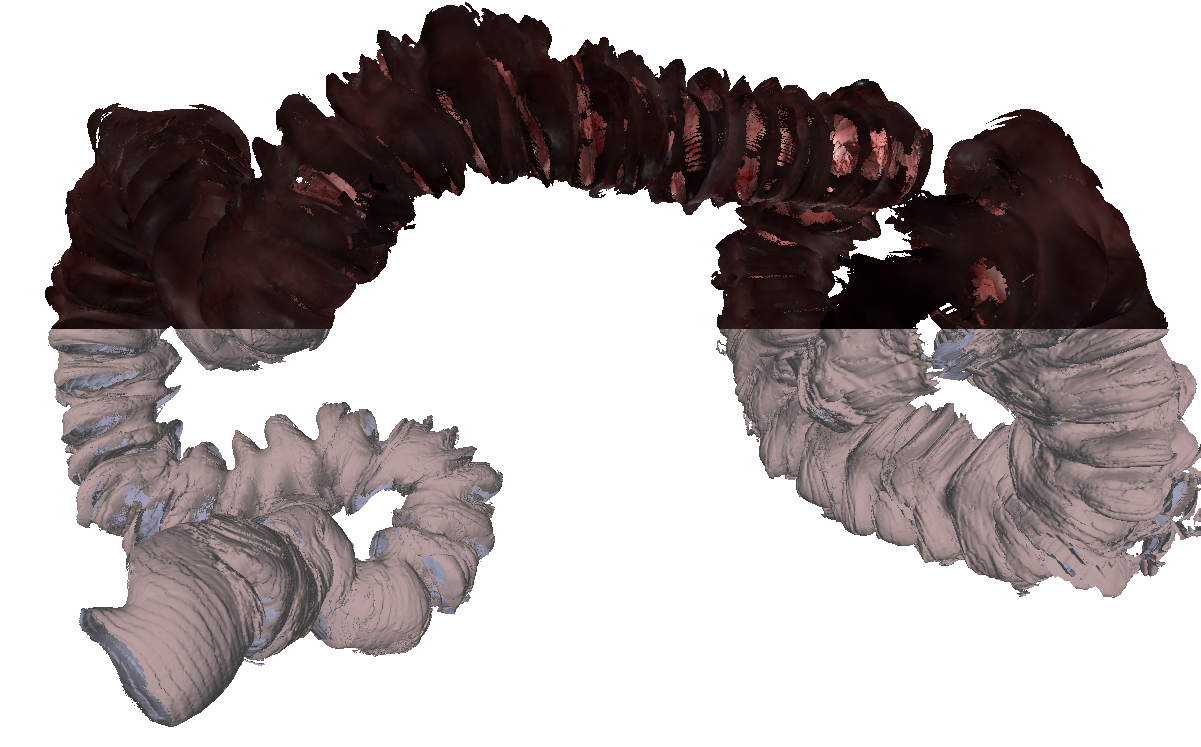}
        \end{subfigure}
        \begin{subfigure}{.12\textwidth}
                \begin{tabular}{c}% if you add [t], than sub images are pushed down
                \smallskip
                    \begin{subfigure}[t]{1\textwidth}
                        \centering
                        \includegraphics[width=1\textwidth]{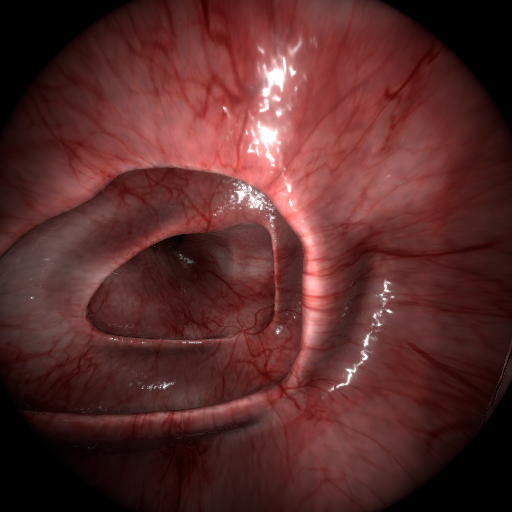}
                    \end{subfigure}\\
                    \begin{subfigure}[t]{1\textwidth}
                        \centering
                        \includegraphics[width=1\textwidth]{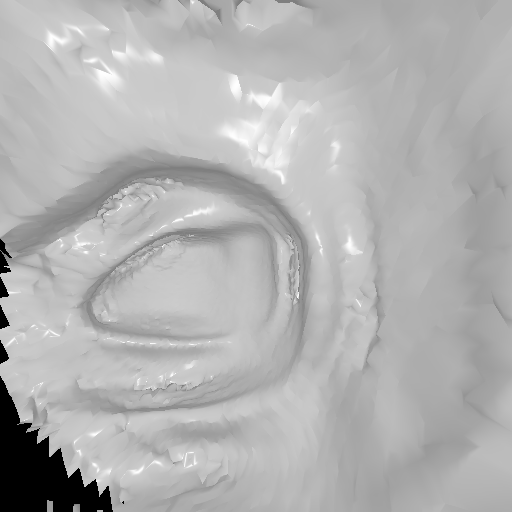}
                    \end{subfigure}
            \end{tabular}
        \end{subfigure}
        \begin{subfigure}{.12\textwidth}
                \begin{tabular}{c}% if you add [t], than sub images are pushed down
                \smallskip
                    \begin{subfigure}[t]{1\textwidth}
                        \centering
                        \includegraphics[width=1\textwidth]{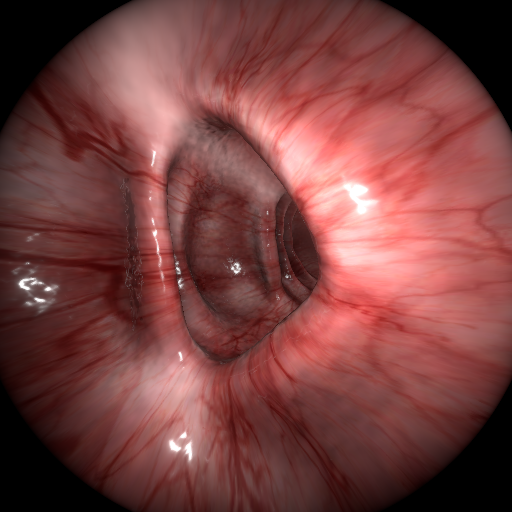}
                    \end{subfigure}\\
                    \begin{subfigure}[t]{1\textwidth}
                        \centering
                        \includegraphics[width=1\textwidth]{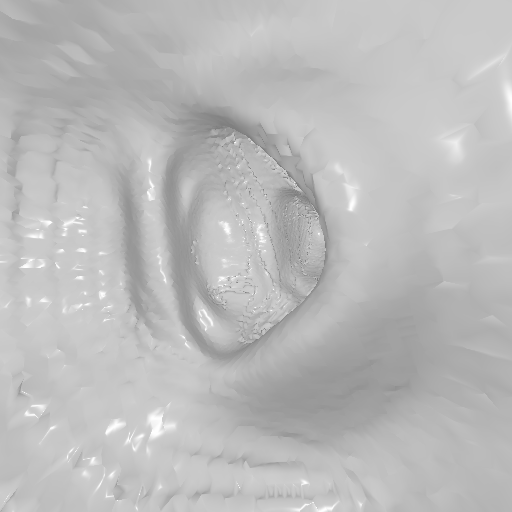}
                    \end{subfigure}
            \end{tabular}
        \end{subfigure}
    \end{tabular}
    % \caption{Left: Full endoscopic colon reconstruction results. Our proposed approach shows high quality scans with negligible camera drift and high local quality of the reconstructions in both geometry and texture. Note that missing region caused by haustral ridge occlusions are visible. Top: the captured RGB images, Bottom: The re-rendered reconstructed model}
    \caption{Left: Full endoscopic colon reconstruction result on the synthetic data-set. Top right: the captured RGB images, Bottom right: The re-rendered reconstructed model}
\label{fig:qualitative_results}
\end{figure}

\begin{figure}[htpb]
    \centering
    \begin{tabular}[t]{ccc}
        \begin{subfigure}{.65\textwidth}
            \centering
            \smallskip
            \includegraphics[width=.8\linewidth]{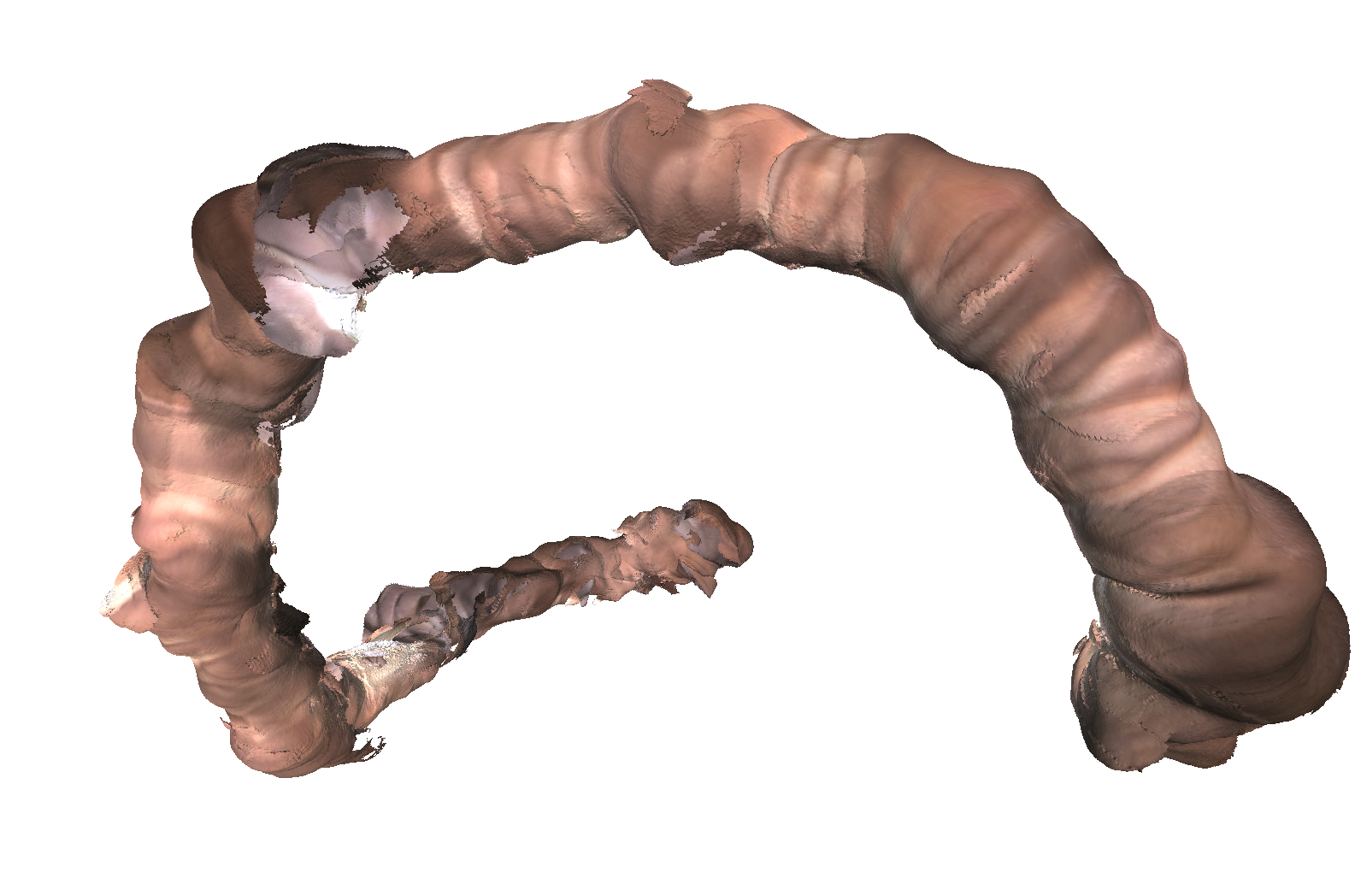}
        \end{subfigure}
        \begin{subfigure}{.35\textwidth}
                \begin{tabular}{c}% if you add [t], than sub images are pushed down
                \smallskip
                    \begin{subfigure}[t]{1\textwidth}
                        \centering
                        \includegraphics[width=.7\textwidth]{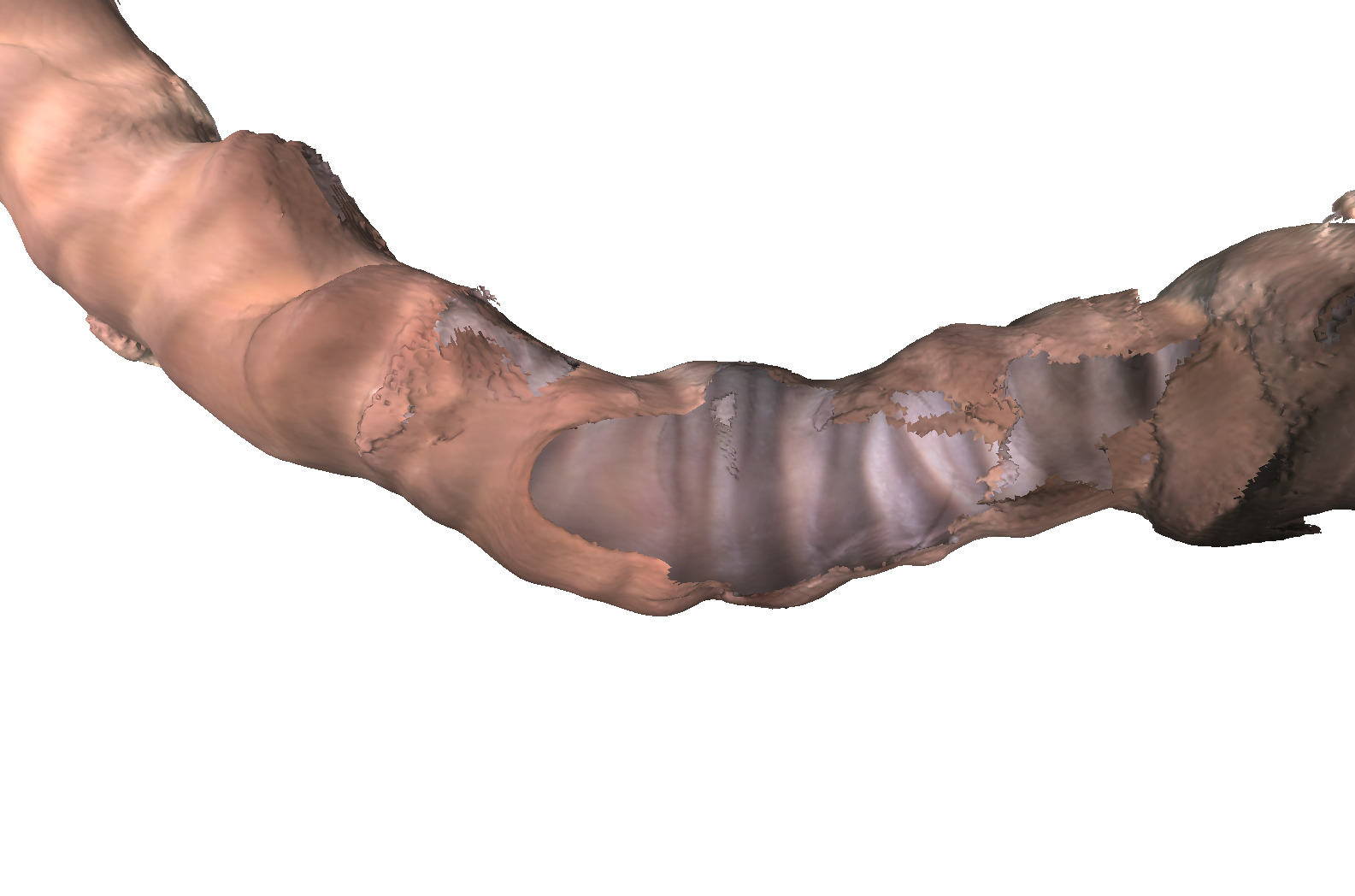}
                    \end{subfigure}\\
                    \begin{subfigure}[t]{1\textwidth}
                        \centering
                        \includegraphics[width=.7\textwidth]{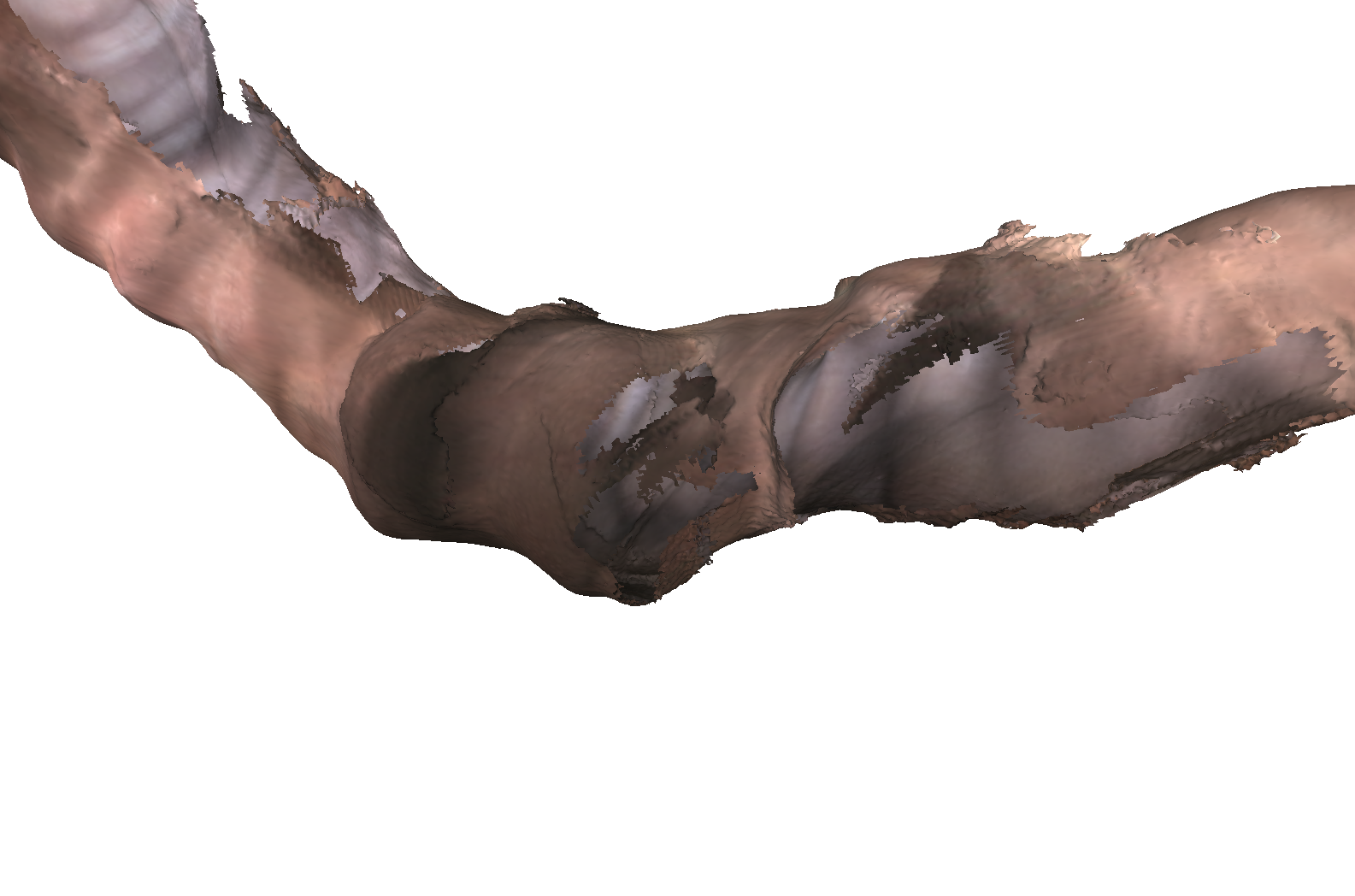}
                    \end{subfigure}
            \end{tabular}
        \end{subfigure}

    \end{tabular}
    \caption{Left: Full endoscopic colon reconstruction result on 3D colon print. Right: zoomed in segments with visible holes (regions that were not covered during the scan)}
\label{fig:qualitative_results_print}
\end{figure}

\begin{figure}[hptb]
    \centering
    \begin{tabular}[t]{ccccc}

       \begin{tabular}{c}% if you add [t], than sub images are pushed down
            %  \parbox[t]{2mm}{\rotatebox[origin=c]{90}{Monodepth2~\cite{monodepth2}}} 
            % \parbox[t]{1mm}{\rotatebox[origin=c]{90}{+ TSDF}} \\\\\\\
            %  \parbox[t]{1mm}{\rotatebox[origin=c]{90}{Ours}}
        \end{tabular}
        
        \begin{subfigure}{.3\textwidth}
                \begin{tabular}{c}% if you add [t], than sub images are pushed down
                \smallskip
                    \begin{subfigure}[t]{1\textwidth}
                        \centering
                        \includegraphics[width=1\textwidth]{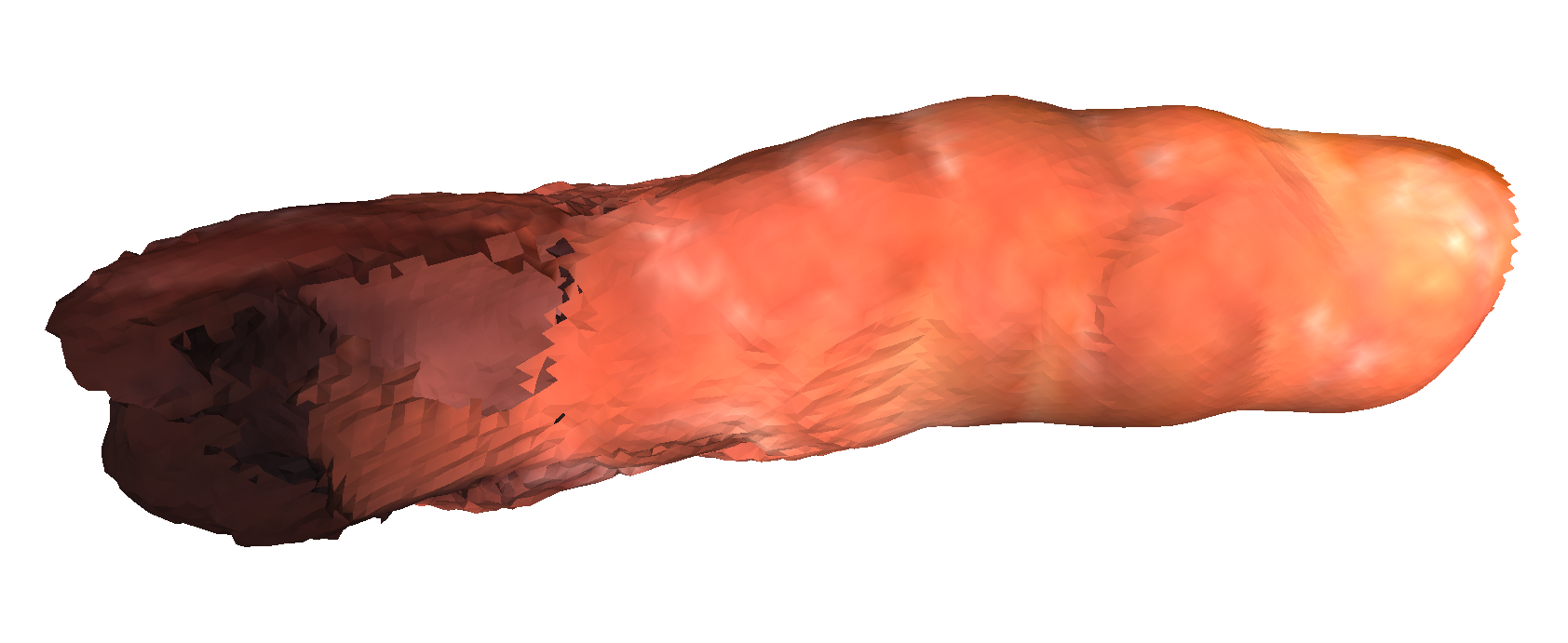}
                    \end{subfigure}\\
                    \begin{subfigure}[t]{1\textwidth}
                        \centering
                        \includegraphics[width=1\textwidth]{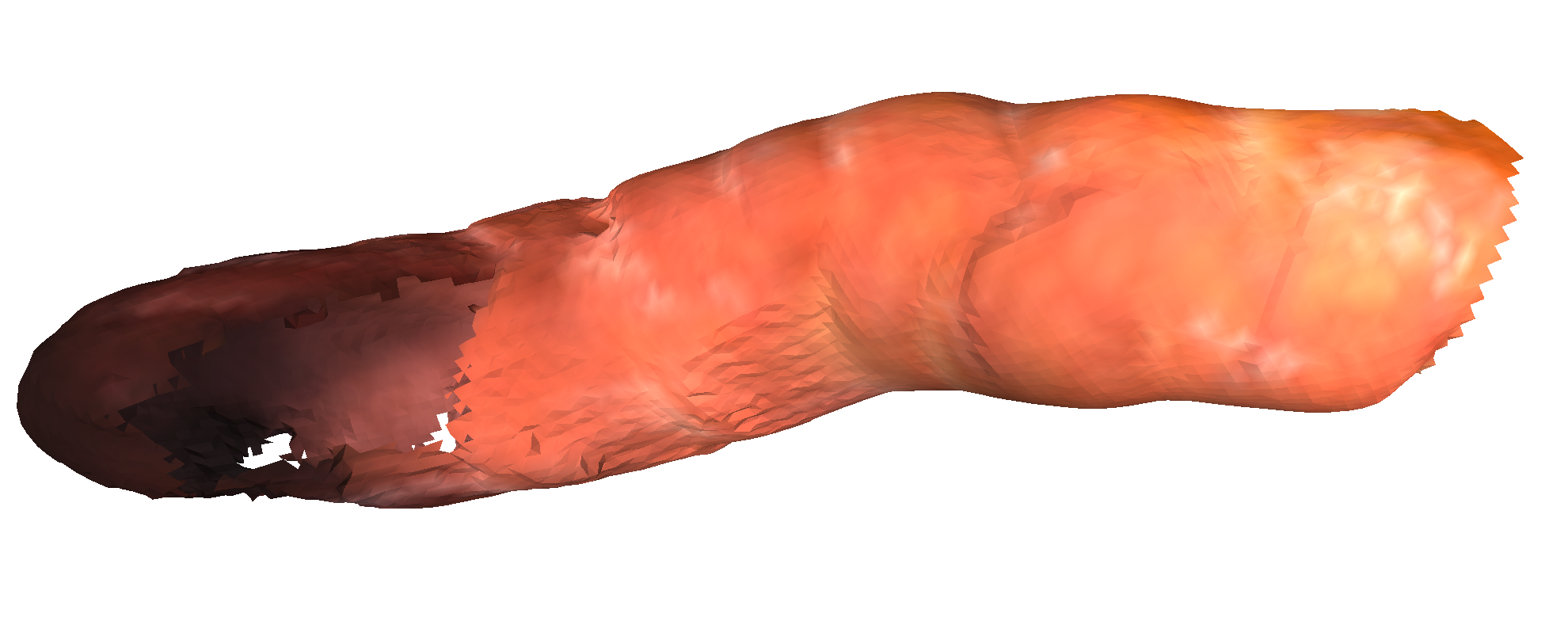}
                    \end{subfigure}\\

            \end{tabular}
            \caption{Seq.02}
             \vspace{-0.2cm}

        \end{subfigure}
        
        \begin{subfigure}{.3\textwidth}
                \begin{tabular}{c}% if you add [t], than sub images are pushed down
                \smallskip
                
                    \begin{subfigure}[t]{1\textwidth}
                        \centering
                        \includegraphics[width=1\textwidth]{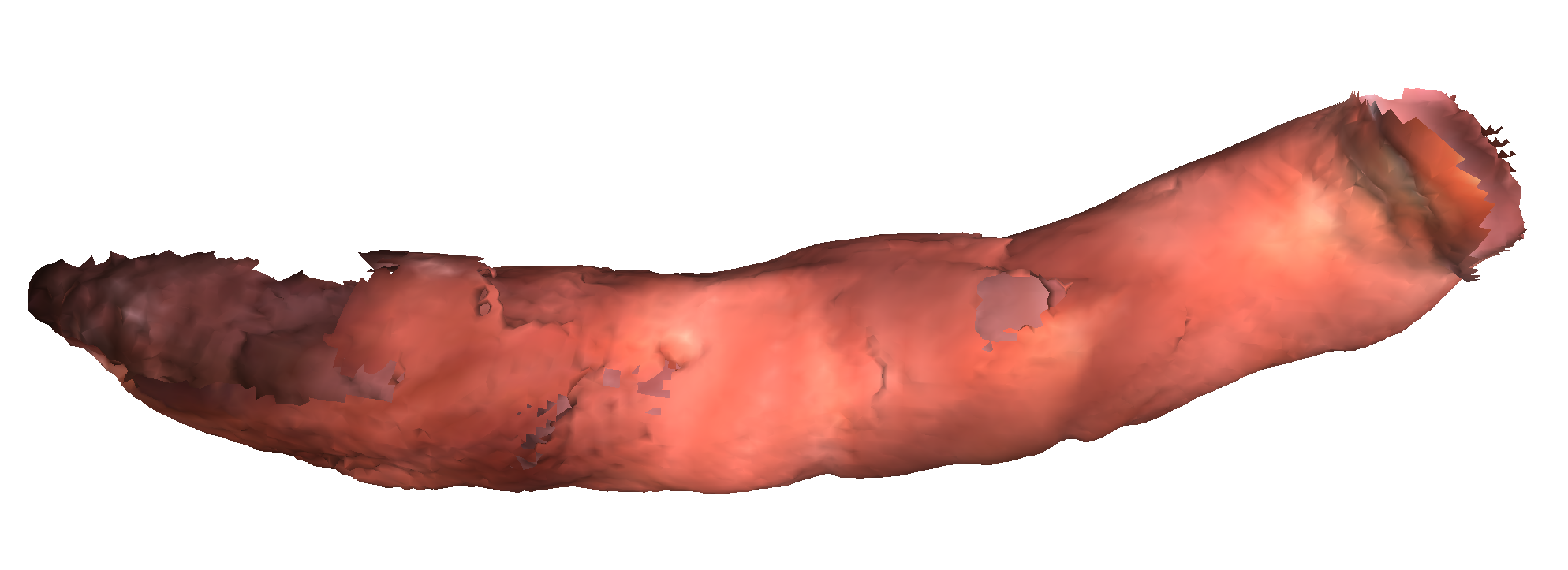}
                    \end{subfigure}\\
                    \begin{subfigure}[t]{1\textwidth}
                        \centering
                        \includegraphics[width=1\textwidth]{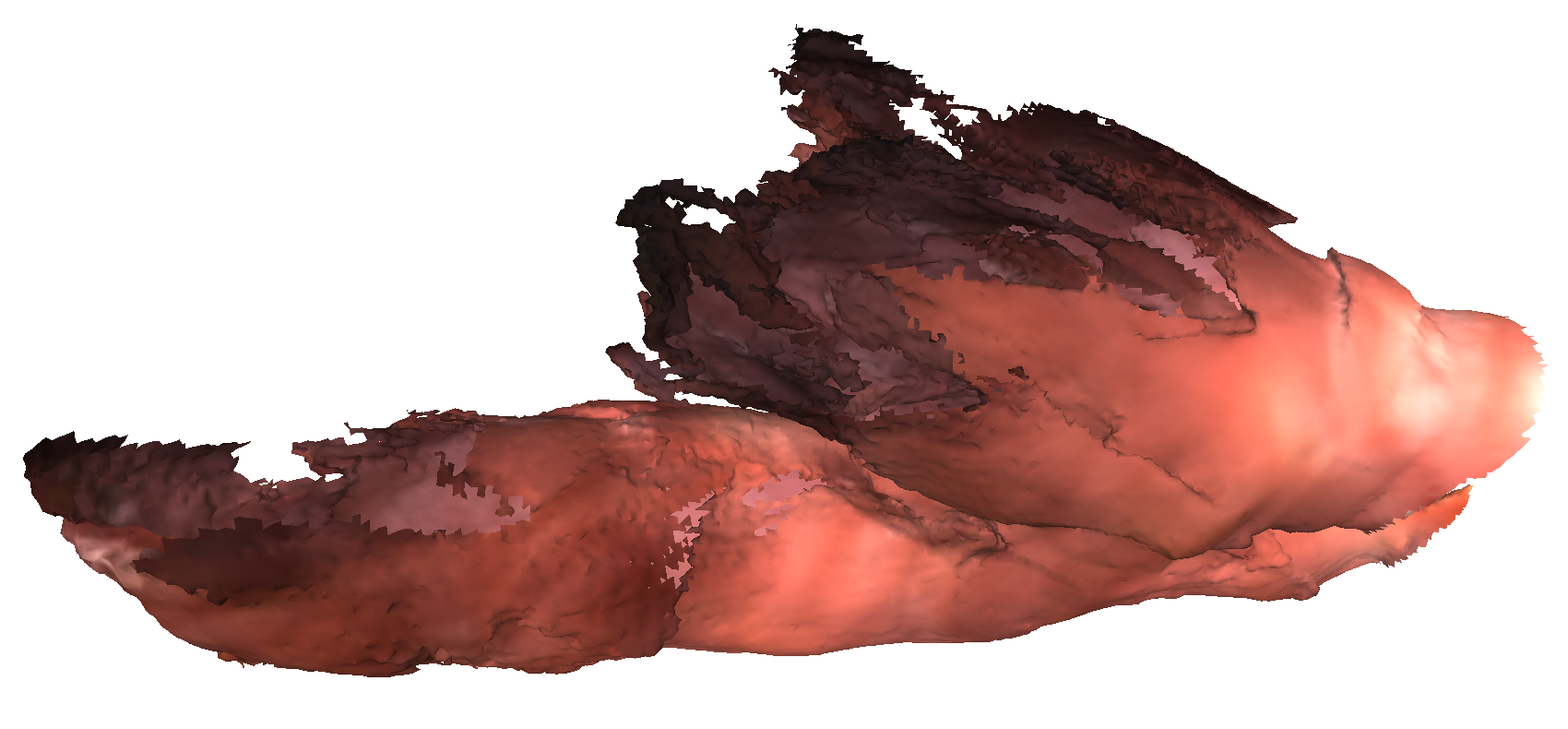}
                    \end{subfigure}\\

            \end{tabular}
            \caption{Seq.09}
             \vspace{-0.2cm}
        \end{subfigure}
        
        \begin{subfigure}{.3\textwidth}
                \begin{tabular}{c}% if you add [t], than sub images are pushed down
                \smallskip
                    \begin{subfigure}[t]{1\textwidth}
                        \centering 
                        \includegraphics[width=1\textwidth]{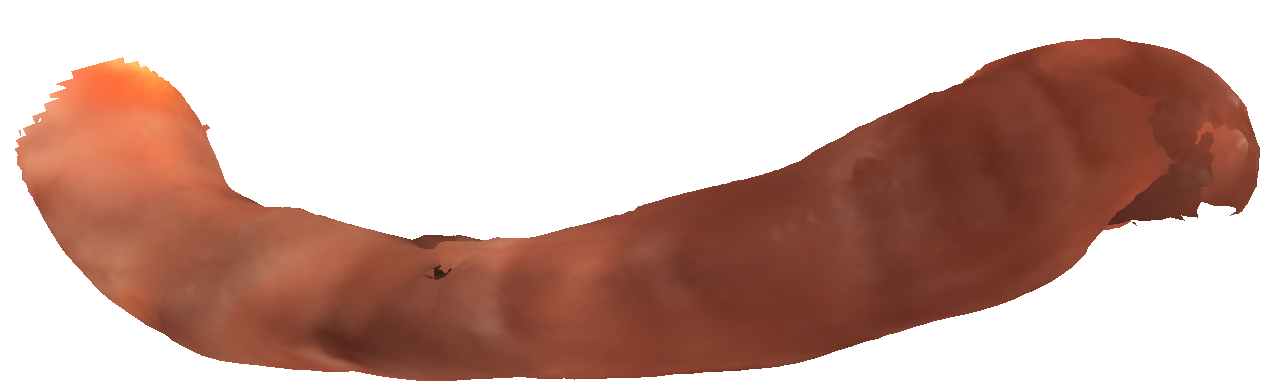}
                    \end{subfigure} \\
                    \begin{subfigure}[t]{1\textwidth}
                        \centering
                        \includegraphics[width=0.9\textwidth]{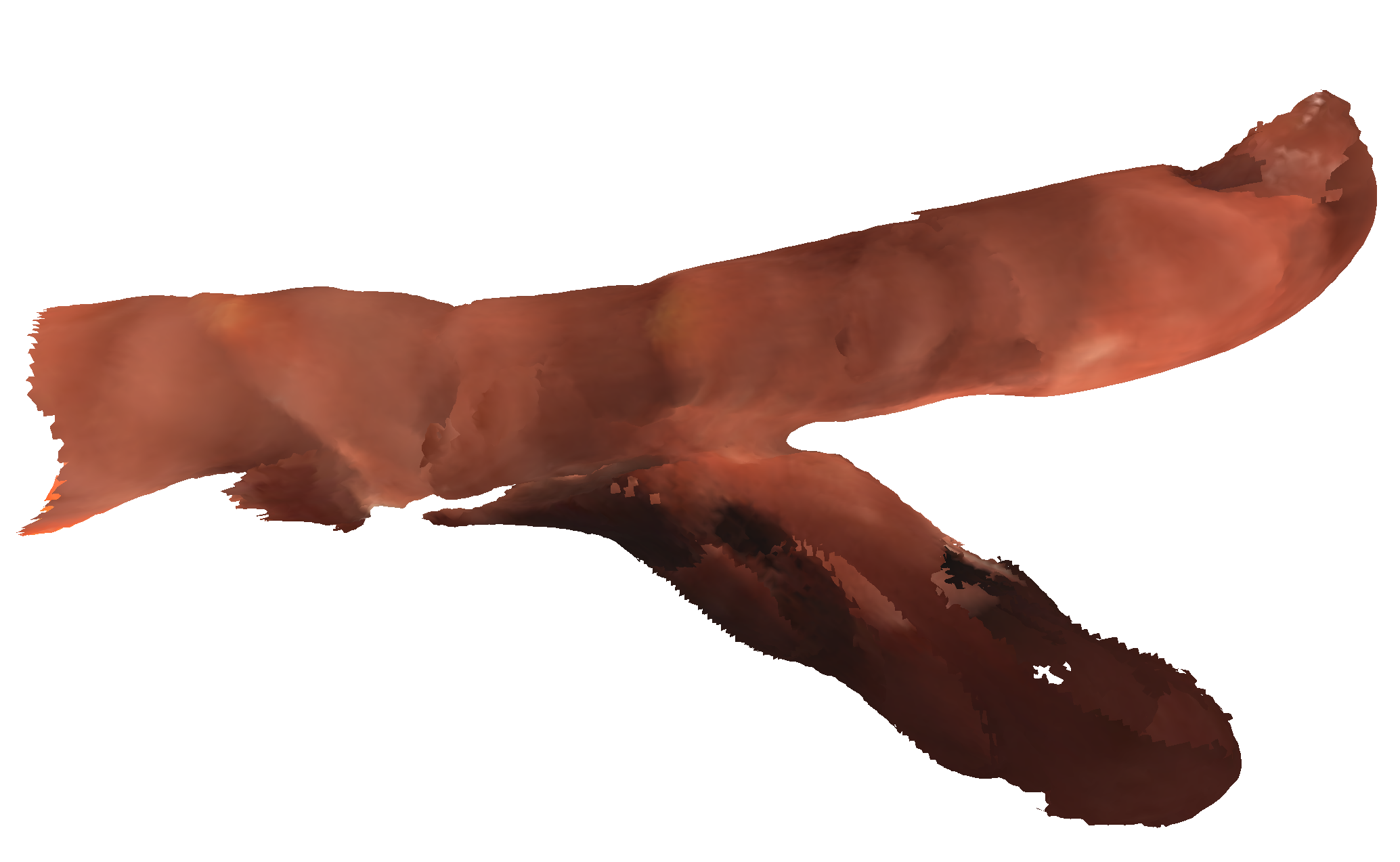}

                    \end{subfigure} \\

            \end{tabular}
            \caption{Seq.18}
        \end{subfigure}

    \end{tabular}
    \caption{3D reconstruction results on Colon10K data-set. Our proposed framework (top) outperforms the mesh reconstructed from depth and pose predictions by Godard et al.~\cite{monodepth2} (bottom)}
\label{fig:reconstruction_qualitative_colon10k}
\end{figure}

\begin{figure}[t]
\includegraphics[width=1\textwidth]{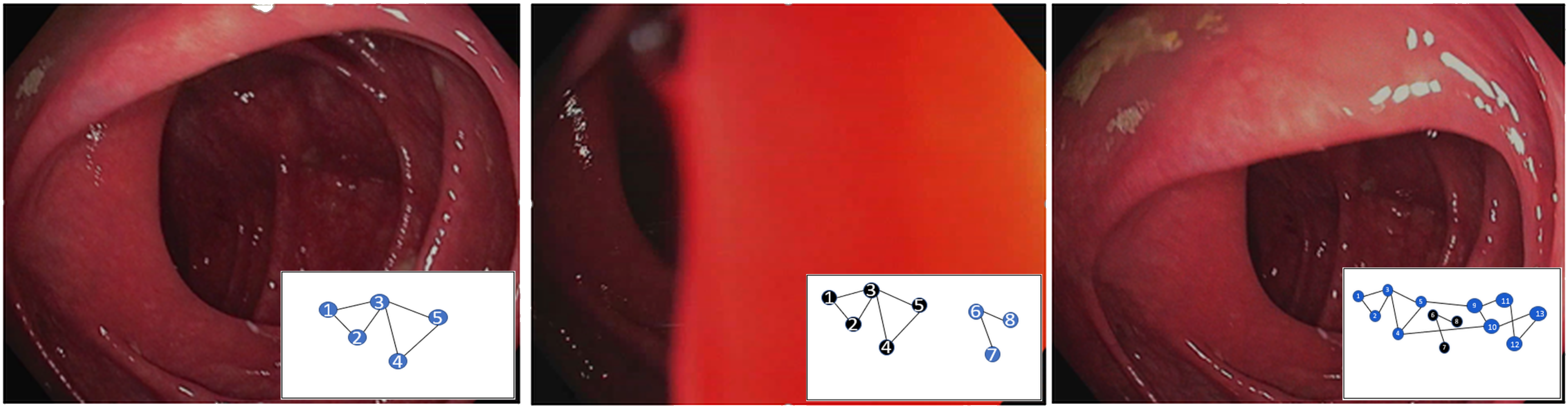}
\caption{Tracking failure recovery on real data: keyframes and their connectivity graph. (middle) recovery is lost (indicated as an additional disjoint sub-graph), (right) recovery is enabled as a connection is formed between frames 5 <-> 9. Active keyframes are highlighted (in blue).}
\label{fig:tracking_failure_recovery}
\end{figure}

\begin{figure}[t]
    \centering
    \begin{tabular}[t]{cc}

        \begin{subfigure}{.45\textwidth}
            \begin{tabular}{c}% if you add [t], than sub images are pushed down
                \smallskip
 
                    \begin{subfigure}[t]{1\textwidth}
                        \centering
                        \includegraphics[width=1\textwidth,height=.45\textwidth]{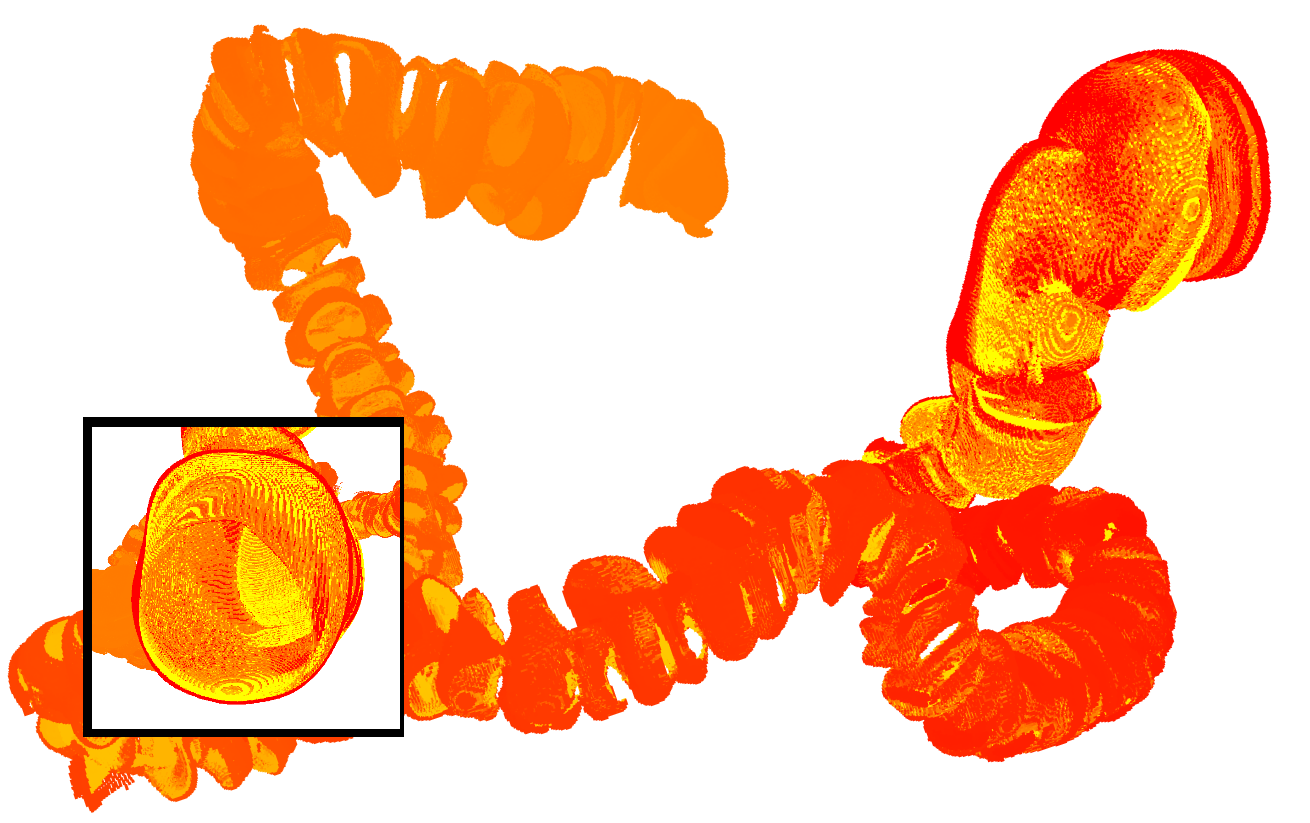}
                    \end{subfigure}
            \end{tabular}
        \end{subfigure}
        \begin{subfigure}{.45\textwidth}
            \begin{tabular}{c}% if you add [t], than sub images are pushed down
                \smallskip

                    \begin{subfigure}[t]{1\textwidth}
                        \centering
                        \includegraphics[width=1\textwidth,height=.45\textwidth]{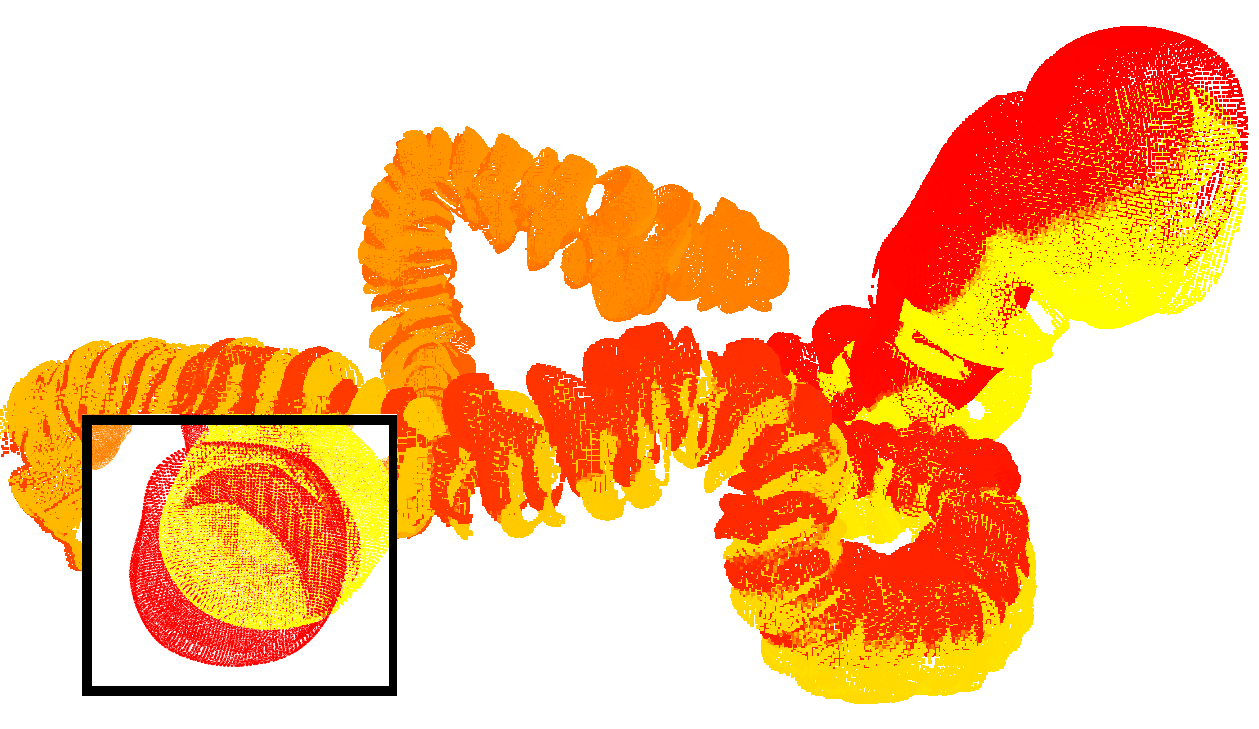}
                    \end{subfigure}
            \end{tabular}
        \end{subfigure}

    \end{tabular}
    \begin{subfigure}[t]{1\textwidth}
        \raggedleft
        \includegraphics[width=.5\textwidth]{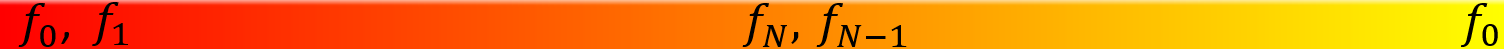}
    \end{subfigure}
    \caption{Comparison of our frame-to-model approach with (left), and without (right) loop closure. The point-clouds are color coded based on their timestamp}
\label{fig:loop_closure}
\end{figure}

\subsection{Quantitative results}
In this section we start by evaluating our framework's components performance. First, our monocular depth estimation is compared to relevant baselines, showing it's strengths in predicting consistent depth-maps. Following with an assessment and comparison of our ContraFeat deep-descriptor in terms of recall/precision. Concluding with dense tracking evaluation, in which we analyze our full pipeline reconstruction trajectory accuracy.
\subsubsection{Monocular depth estimation}
We train our network and~\cite{monodepth2} on the synthetic data-set with the same data split paradigm as described above and compare the results.
To accommodate different scales of depth-maps, we use per-image median ground-truth scaling as introduced by~\cite{DBLP:journals/corr/ZhouBSL17}. As shown in Table~\ref{tab:depth_pred_quantitative}, our monocular method outperforms existing state-of-the-art self-supervised approaches. Specifically, our Sq Rel error has improved substantially (15.5\% relatively) over~\cite{monodepth2} which corresponds to improved depth consistency. 

\begin{table}[tbp]
    \caption{Quantitative results on the synthetic data}
    \label{tab:depth_pred_quantitative}
    \centering
        \begin{tabular}{@{}lccccccc@{}}
        \toprule
        Method & Abs Rel  & Sq Rel & RMSE & RMSE log  & $\delta<1.25$ & $\delta<{1.25}^2$ & $\delta<{1.25}^3$  \\ 
        \midrule
        
        MonoDepth2~\cite{monodepth2} & 0.089 & 0.058 & 0.540 & 0.134 & 0.931 & \textbf{0.982} &      0.992  \\ 
        ColDE \cite{zhang2021colde}  \textdagger & 0.077 &  0.079 &  0.701 &  0.134 &  0.935 &  0.975 &  0.989 \\ 
        \textbf{Ours}   &  \textbf{ 0.075} & \textbf{0.049} & \textbf{0.521} & \textbf{0.126} & \textbf{0.94} & 0.980 & \textbf{0.992}  \\
        \bottomrule
        \end{tabular}
        \begin{tablenotes}
        \small
        \item \textdagger Results taken from paper since training data-set and code are not publicly available.
        \end{tablenotes}

\end{table}

\subsubsection{Keypoints correspondence evaluation}

We estimate the keypoint matching performance as precision (i.e., the percentage of \textbf{correct} correspondence from all correspondences found) and recall (i.e., the percentage of valid correspondences from all ground-truth correspondences) on the synthetic data-set with ground truth depth and pose. To determine if a keypoint pair correspondence is correct, we test if it lies in the ground truth set. We evaluate our contrastive deep-feature (\textit{CDF}) performance by comparing it to SIFT-based and SuperPoint~\cite{super_point}, with and without the addition of the key points filter (\textit{kpf}) described in Sec~.\ref{feature_corr}. Table~\ref{tab:loop_closure_pre_recall} shows that SIFT matches suffers from a low recall and extremely low precision which can impact the overall reconstruction. The additional \textit{kpf} increases the precision drastically to around $99.5\%$ on average, at the cost of lowering the recall as can be expected. Although SuperPoint matches start with higher precision/recall than SIFT, after applying the key points filter, the precision greatly improves at the cost of drastically decreasing the recall. This could indicate that the localization error of the matched keypoints is high and therefore get filtered out. The \textit{CDF} matches starts with much higher recall $\&$ precision, combined with \textit{kpf} it achieves outstanding results of $74.9\%$ recall and $100\%$ precision on average. The results show that our hybrid approach i.e., using SIFT keypoint with contrastive deep-features, is well suited to OC domain.
See appendix ~\ref{corr_qualitative_res} for examples of predicted correspondences between frames.

\begin{table}[htbp]
\begin{minipage}{1\textwidth}
\centering
        \caption{Keypoints recall/precision on the synthetic data-set: SIFT, SuperPoint (SP) and our ContraFeat deep feature (CDF) with/without our key points filter (kpf)}
        \label{tab:loop_closure_pre_recall}

        \begin{tabular}{@{}lccccccl@{}}\toprule
         & Seq. & SIFT$[\%]$& SIFT+kpf$[\%]$ & SP$[\%]$ &SP+kpf$[\%]$ & CDF$[\%]$& CDF+kpf$[\%]$\\ \midrule
         & Seq1 & 31.4/12.1 & 24.2/100   & 62.4/78.9 & 22.4/100 & 74.2/86.1 & \textbf{70.2/100} \\
         & Seq2 & 44.4/32.1 & 29.9/98.4  & 59.2/68.2 & 17.3/100 &  82.3/89.2 & \textbf{79.2/100} \\  
         & Seq3 & 37.8/21.3 & 26.8/100   & 55.4/64.8 & 16.2/100 &  78.1/83.5 & \textbf{75.2/100} \\
         \bottomrule
        \end{tabular}
\end{minipage}
\end{table}
\subsubsection{Full endoscopic reconstruction}

We evaluate the trajectory accuracy over the synthetic data-set and the 3D colon print sequence, and compare it to Direct Sparse Odometry (DSO)~\cite{DSO}. We also provide results for the SIFT-based feature descriptors instead of our suggested deep feature descriptor.~Table~\ref{tab:trajectory_accuracy} summarizes the ATE~\cite{ATE} results. As can be seen from the rmse and std values, our method surpasses DSO even using SIFT-based feature descriptors. When incorporating the deep features, our final approach is $46\%$ better than DSO on the synthetic datast and $79\%$ on the 3D colon print. This shows the major contribution of our framework to the reconstruction. It is important to note that, although ContraFeat was only trained on synthetic data, the ATE has improved by $10\%$ relatively over SIFT on the 3D print model. Thus, demonstrating ContraFeat's robustness. See appendix~\ref{trajectories_comparison} for visual trajectories comparison.

\begin{table}[htbp]
\small
\begin{minipage}{1\textwidth}
\centering
        \caption{Avg. ATE statistics over synthetic and 3D colon print data-sets (avg. trajectory of 125cm)
        }
        \label{tab:trajectory_accuracy}
        \begin{tabular}{@{}lcccc@{}} \toprule
        \multicolumn{1}{l}{Method}
          & \multicolumn{2}{c}{synthetic data-set} & \multicolumn{2}{l}{3D colon print data-set} \\ \cmidrule{2-5}
         &
          \multicolumn{1}{c}{RMSE{[}cm{]}} &
          std{[}cm{]} &
          \multicolumn{1}{c}{RMSE{[}cm{]}} &
          std{[}cm{]} \\ \midrule
        \multicolumn{1}{l}{DSO}        & \multicolumn{1}{c}{53.4}     & 12.3     & \multicolumn{1}{c}{37.1}             & 13.9            \\ 
        \multicolumn{1}{l}{Ours(SIFT)} & \multicolumn{1}{c}{36.1}     & 3.3      & \multicolumn{1}{c}{8.72}     & \textbf{3.53}           \\         \multicolumn{1}{l}{Ours}       & \multicolumn{1}{c}{\textbf{28.9}}     & \textbf{2.7}      & \multicolumn{1}{c}{\textbf{7.88}}     &   3.75              \\ 
        \bottomrule
        \end{tabular} 
\end{minipage}\hfill % maximize space between the minipages
\end{table}

\subsection{Qualitative results} \label{qual_res}
\subsubsection{Monocular depth estimation}

In Fig.~\ref{fig:monocular_depth_prediction_qualitative} we qualitatively compare our predicted depth maps with monodepth2~\cite{monodepth2}. Note that our method produces high quality depth-maps characterized with consistent depth around the colon's surface while maintaining sharp boundaries around haustral folds. Furthermore, our method is more robust to specular reflection artifacts. Extra depth-maps can be seen in appendix~\ref{A. extra ct dm}.

\subsubsection{Full endoscopic reconstruction}

Fig.~\ref{fig:qualitative_results} and Fig.~\ref{fig:qualitative_results_print} depict the qualitative results of our method; showing the reconstructions of a fully endoscopic colon investigation on the synthetic data-set and the 3D colon print respectively. Fig.~\ref{fig:qualitative_results} demonstrates that our approach produces high quality scans with negligible camera drift and high local quality of the reconstructions in both geometry and texture. Note in Fig.~\ref{fig:qualitative_results}, the clear resemblance between the re-rendered mesh of the reconstructed model (in gray) and the captured RGB images. We are able to successfully capture the geometric curvatures of the colon while keeping the missing regions visible.
In addition, we show in Fig.~\ref{fig:reconstruction_qualitative_colon10k} a qualitative comparison of the reconstructed surfaces based on real optical colonoscopy videos from Colon10K.
The results from MonoDepth2~\cite{monodepth2} 
were produced using their pre-trained network on ImageNet, which was fine-tuned with semi-supervision over the data-set. To be comparable, we additionally fused~\cite{monodepth2} outputs as described in Sec.\ref{3d_reco} in order to generate a mesh. Extra images from different point of views are shown in appendix~\ref{A. extra 3d qual}.

Our novel hierarchical global pose optimization framework implicitly handles loop closure, recovers from tracking failures, and reduces geometric drift. Our method is able to support multiple loop closure as it does not rely on explicit loop closure detection; thus, scales better.

\paragraph{Tracking failure recovery}
When a new keyframe cannot be aligned with any of the previous keyframes, tracking is assumed to be lost. Sensibly, this means that the keyframe won't have any edges connecting it to any previous keyframes in the pose-graph optimization. Thus, an additional lone fragment will be created in the fragment's connectivity graph and the predicted surface won't be included in the output reconstruction. Based on our approach, recovery is enabled to any previously scanned areas as we don't require temporal nor spatial coherence. 
A common tracking loss sequence is shown in Fig.~\ref{fig:tracking_failure_recovery} in which the camera is occluded by a Haustral fold (colon wall protrusions). As our method globally matches new keyframes against all existing keyframes, tracking can be lost and recovered at a completely different place. 

% As can be seen in Fig.~\ref{fig:tracking_failure_recovery}, the resulted connectivity graph from the optimization process shows that re-localization is enabled as a connection is formed between frames 5 <-> 10 and consequently, recovery is enabled.
\paragraph{Loop closure operation}
Our global pose optimization continuously operates in the background; detects and handle loop closures seamlessly thus mitigating camera pose errors and evidently preventing geometric drifts over time. See Fig.~\ref{fig:loop_closure} where the synthetic sequence is played forward and backward in order to create a loop. Notice how the forward and backwards trajectories align in our method versus the vanilla frame-to-model approach.
\section{Conclusion} \label{conclusion}
We have presented a novel deep learning 3D reconstruction approach that provides a robust tracking with negligible geometric-drift and implicitly solves the tracking loss problem frequent during OC. The proposed approach was evaluated on multiple data-sets, showing outstanding reconstruction quality and completeness compared to previously suggested methods. Additional experiments were conducted to illustrate the proficiency of the suggested method in several difficult cases common in colonoscopic sequences not supported by previous methods.
The reconstruction can be used to indicate un-inspected surfaces that could contain colorectal lesions and decrease the miss rates of polyps. Although our framework deals with real life issues common in OC, some still remains unattended: non-rigidity of the colon can be suppressed within the TSDF only up to some extent and dynamic objects like stool and degenerate frames needs to be discarded. We leave these further explorations
to future work.

% \begin{ack}
% Use unnumbered first level headings for the acknowledgments. All acknowledgments
% go at the end of the paper before the list of references. Moreover, you are required to declare
% funding (financial activities supporting the submitted work) and competing interests (related financial activities outside the submitted work).
% More information about this disclosure can be found at: \url{https://neurips.cc/Conferences/2022/PaperInformation/FundingDisclosure}.

% Do {\bf not} include this section in the anonymized submission, only in the final paper. You can use the \texttt{ack} environment provided in the style file to autmoatically hide this section in the anonymized submission.

% \end{ack}

\bibliography{bib.bib}

\clearpage
\pagebreak 

\appendix

\section{Appendix}

\subsection{Synthetic data generation}\label{A. synthetic_data_generation}
For the purpose of reproducibility, we state the parameters that were used to build each synthetic sequence using the synthetic colonoscopy simulator~\cite{UTS}. The parameters are summarised in Table.~\ref{tab:data_creation}, where \textit{RP} stands for \textit{Random Path}. It is worth mentioning that the user does not have the ability to set the seed of the random number generator for the random path chosen.

\subsection{Depth Training and implementation details}\label{A. imple_details}
We use AdamW optimizer~\cite{Loshchilov2019DecoupledWD}, with $\beta_1 = 0.9$,
$\beta_2 = 0.999$. We train the synthetic and the Colon10k models for 40 epochs, with a batch
size of 16 on a 24GB Nvidia 3090 RTX. The initial learning rate is
$10^{-4}$; we reduce it by half on each of the 16th, 24th and 32nd epochs.
As for the 3D colon print model, we train for 200 epochs; reduce the learning rate by half on each of the 80th, 120th and 170th epochs.
We center-crop the synthetic images to 400×400 to remove vignetting effects. The Colon10k images are provided in an un-distorted and center-cropped version of 270x216 pixels. Finally, the cropped image is scaled to 224x224 before feeding to the network. For the 3D colon print, we employ test time training due to the scarcity of the data and the fact that the training process is completely self-supervised.

To generate the specular reflection mask for each frame, we convert the input frames to YUV color-space and apply a threshold of 90\% on the Y channel and dilate the resulting binary mask with a kernel of 13 pixels.

We use MMLab's \cite{MMDetection} implementation of ResNet~\cite{resnet}, deformable convolutions and FPN. All ResNet encoders and the FPN were pre-trained on ImageNet~\cite{ImageNet}.
We use ResNet50 for the depth encoder. For the pose encoder and FPN, we use ResNet18. Deformable convolution layers are applied in the depth encoder stages of conv3, conv4 and conv5. We set $\lambda_{ph-extra}=0.1$ and $\lambda_{dc}=0.1$, $\tau=0.01$.

\subsection{Correspondence matching qualitative results}
\label{corr_qualitative_res}
\begin{figure}[htb]
    \centering
    \begin{tabular}[t]{ccc}
    SIFT~\cite{Lowe:2004uq} & SuperPoint~\cite{super_point} & ContraFeat \\

        % \end{tabular}
        \begin{subfigure}{.3\textwidth}
                \begin{tabular}{c}% if you add [t], than sub images are pushed down
                    \begin{subfigure}[t]{1\textwidth}
                        \centering
                       \includegraphics[width=1\textwidth]{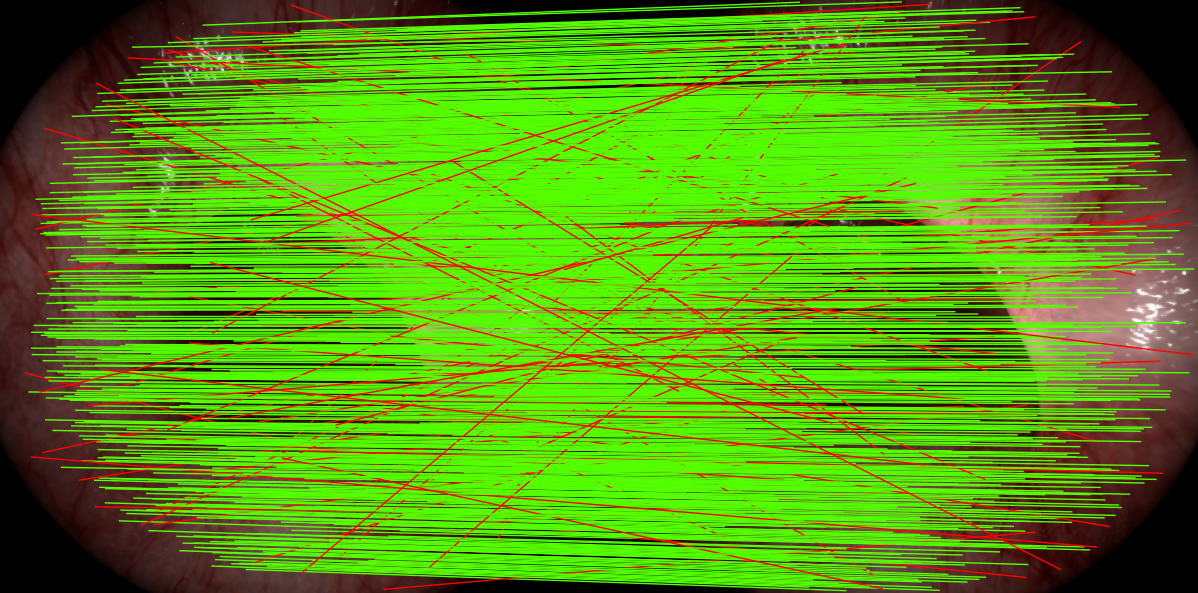}
                    \end{subfigure}\\
                    \begin{subfigure}[t]{1\textwidth}
                        \centering
                        \includegraphics[width=1\textwidth]{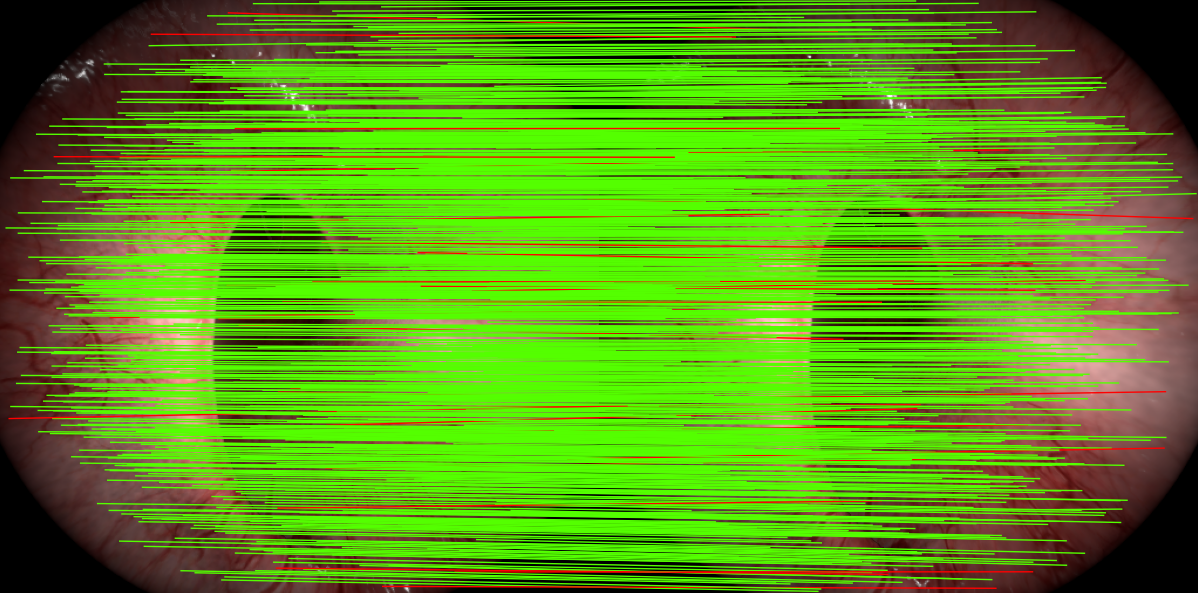}
                    \end{subfigure}\\
                    \begin{subfigure}[t]{1\textwidth}
                        \centering
                       \includegraphics[width=1\textwidth]{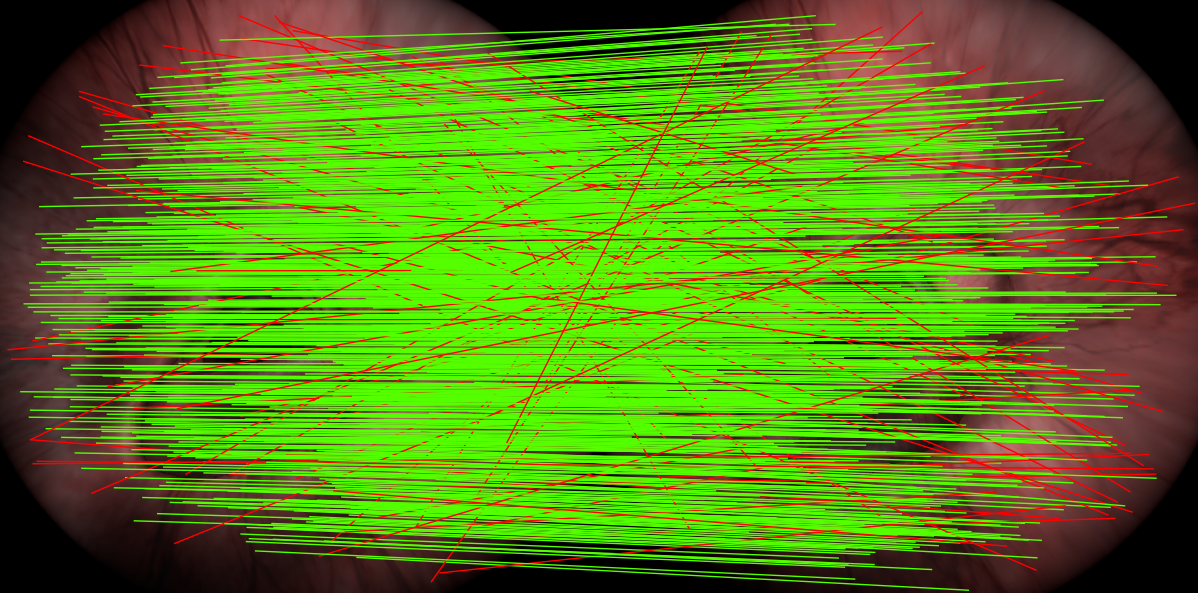}
                    \end{subfigure}\\

            \end{tabular}
        \end{subfigure} &
        \begin{subfigure}{.3\textwidth}
                \begin{tabular}{c}% if you add [t], than sub images are pushed down
                    \begin{subfigure}[t]{1\textwidth}
                        \centering
                        \includegraphics[width=1\textwidth]{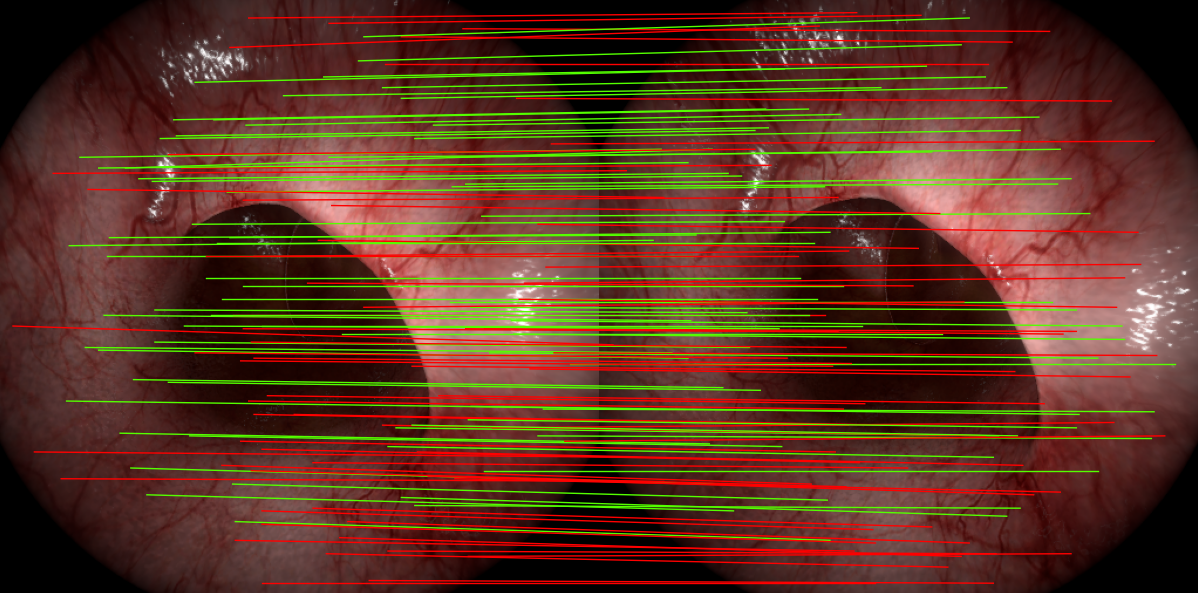}
                    \end{subfigure}\\
                    \begin{subfigure}[t]{1\textwidth}
                        \centering
                        \includegraphics[width=1\textwidth]{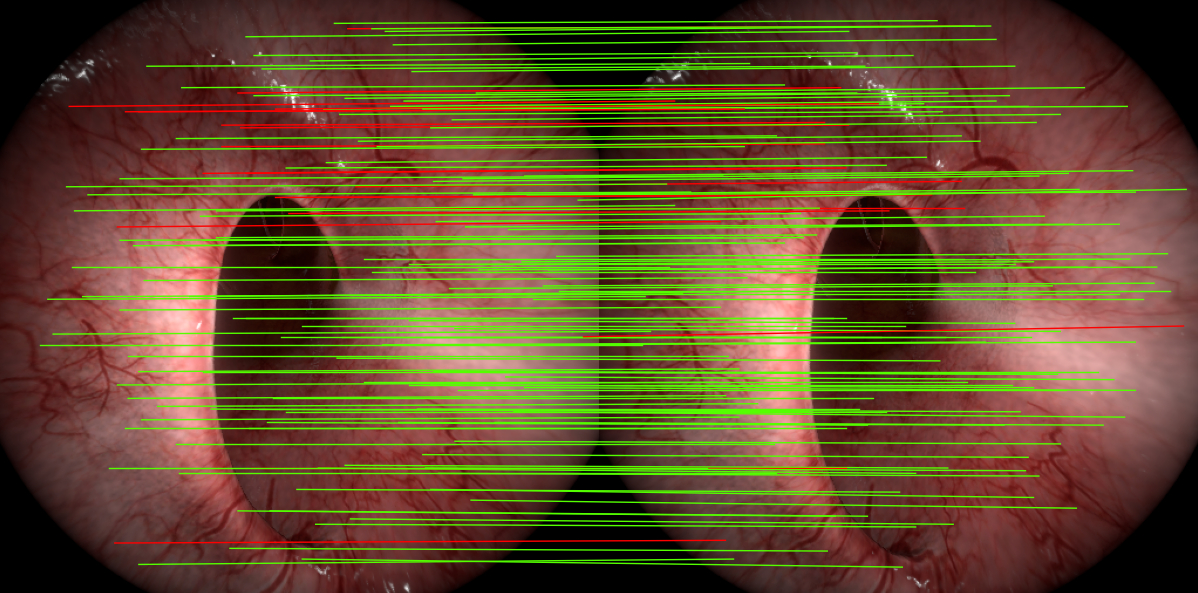}
                    \end{subfigure}\\
                    \begin{subfigure}[t]{1\textwidth}
                        \centering
                        \includegraphics[width=1\textwidth]{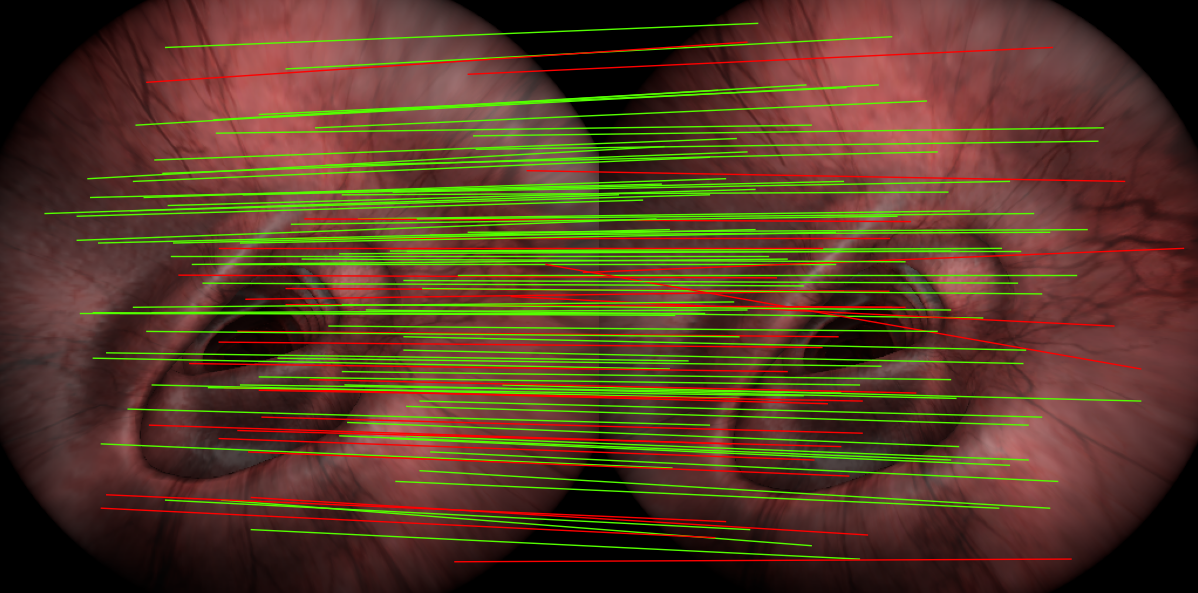}
                    \end{subfigure}\\

            \end{tabular}
        \end{subfigure} &
        \begin{subfigure}{.3\textwidth}
                \begin{tabular}{c}% if you add [t], than sub images are pushed down
                    \begin{subfigure}[t]{1\textwidth}
                        \centering
                        \includegraphics[width=1\textwidth]{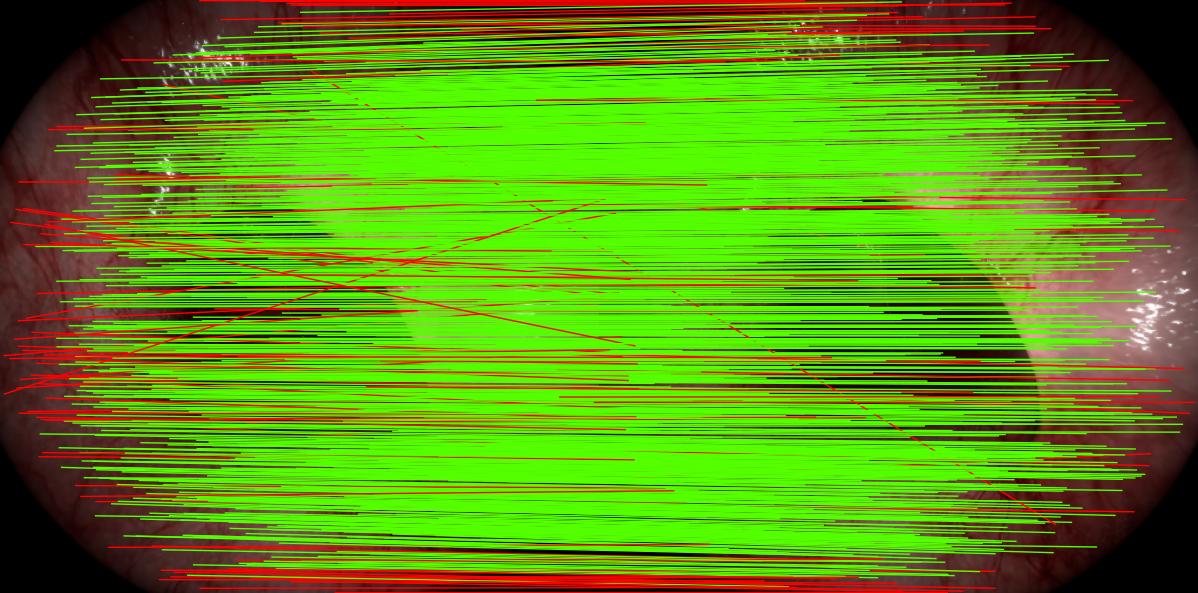}
                    \end{subfigure}\\
                    \begin{subfigure}[t]{1\textwidth}
                        \centering
                        \includegraphics[width=1\textwidth]{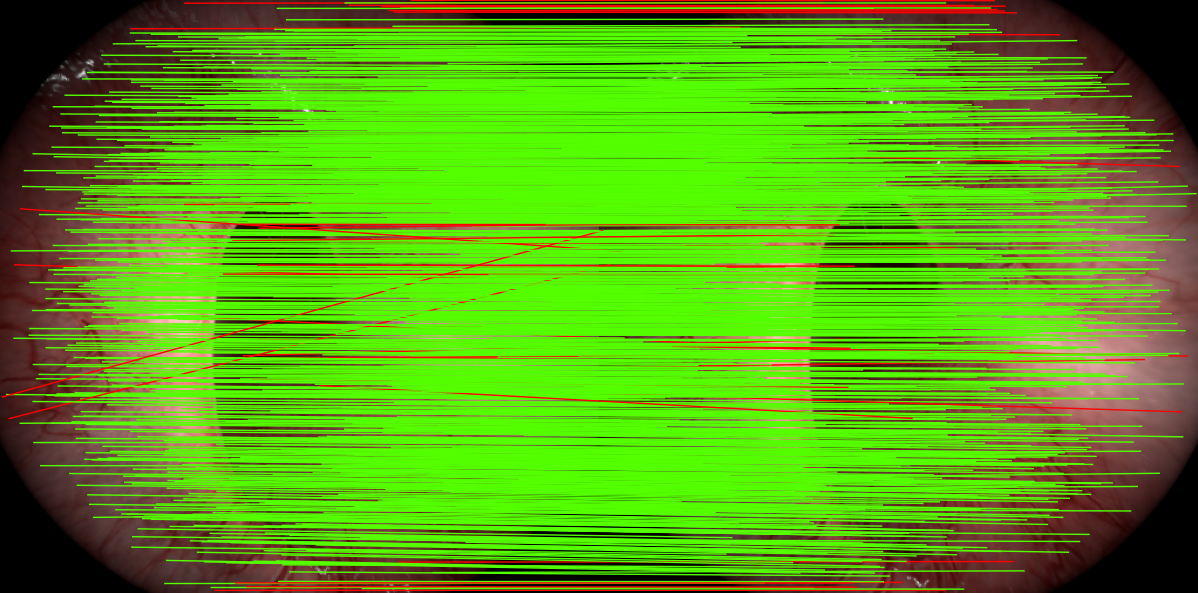}
                    \end{subfigure}\\
                    \begin{subfigure}[t]{1\textwidth}
                        \centering
                       \includegraphics[width=1\textwidth]{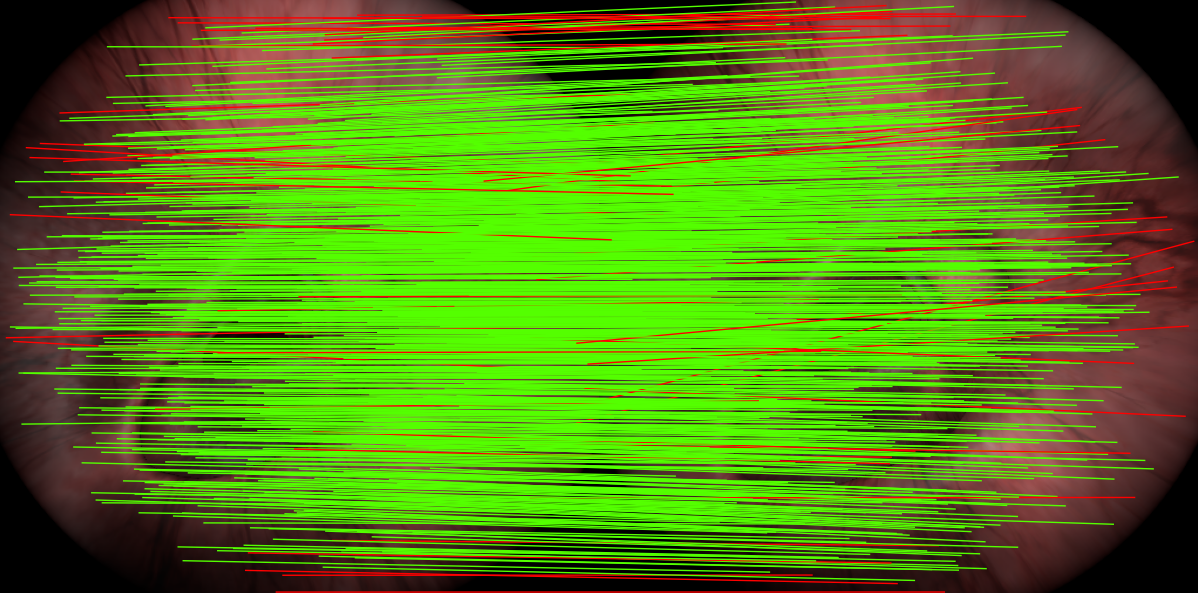}
                    \end{subfigure}\\

            \end{tabular}
        \end{subfigure} 
    \end{tabular}
    \caption{Matching qualitative comparison on the synthetic data. Correct matches are \textcolor{green}{green} lines and mismatches are \textcolor{red}{red} lines. Mismatches defined when correspondence re-projection error is greater than 1$\%$ of colons diameter.}
\label{fig:features_qualitative_comparison_uts}
\end{figure}

Matching examples of ContraFeat,  SIFT~\cite{Lowe:2004uq} and SuperPoint~\cite{super_point} are shown in Fig.~\ref{fig:features_qualitative_comparison_uts}. ContraFeat incline to produce more correct matches and spread out evenly throughout the image, and is less susceptible against drastic illumination changes. 

\subsection{Comparison of the estimated trajectories and ground truth trajectories}
\label{trajectories_comparison}
\begin{figure}[htpb]
    \begin{tabular}[t]{ccc}
        DSO~\cite{DSO} & Ours + SuperPoint~\cite{super_point} & Ours \\

        \begin{subfigure}{.32\textwidth}
                \begin{tabular}{c}% if you add [t], than sub images are pushed down
                    \begin{subfigure}[t]{1\textwidth}
                       \includegraphics[width=1\textwidth]{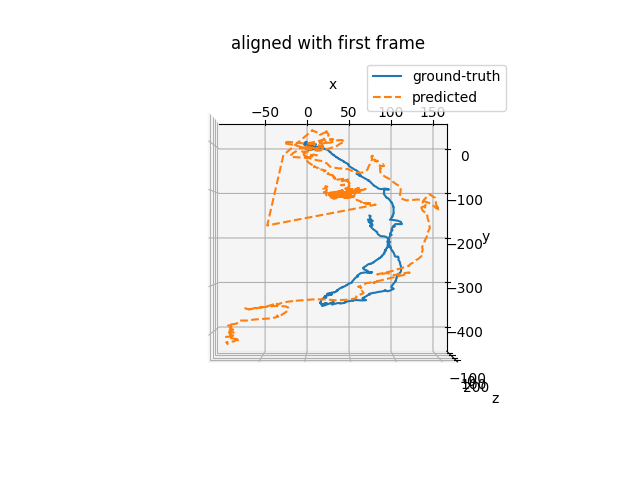}
                    \end{subfigure}\\

            \end{tabular}
        \end{subfigure} &
                \begin{subfigure}{.32\textwidth}
                \begin{tabular}{c}% if you add [t], than sub images are pushed down
                    \begin{subfigure}[t]{1\textwidth}
                       \includegraphics[width=1\textwidth]{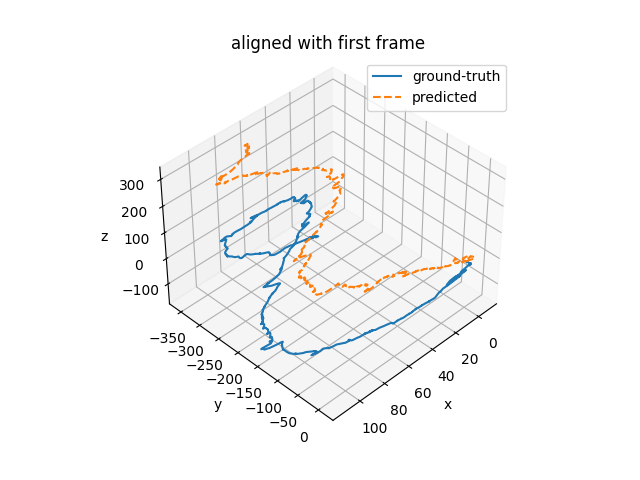}
                    \end{subfigure}\\

            \end{tabular}
        \end{subfigure} &

        \begin{subfigure}{.32\textwidth}
                \begin{tabular}{c}% if you add [t], than sub images are pushed down
                    \begin{subfigure}[t]{1\textwidth}
                        \includegraphics[width=1\textwidth]{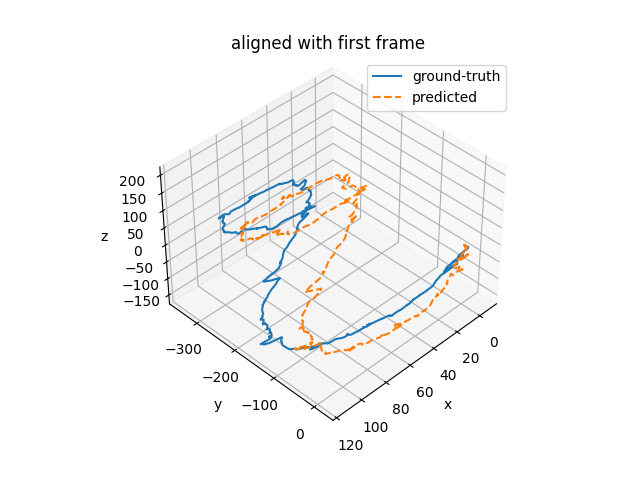}
                    \end{subfigure}\\

            \end{tabular}
        \end{subfigure} 

    \end{tabular}
    \caption{Comparison of the estimated trajectories and ground truth trajectories on the 3D colon print sequence}
\label{fig:quality_traj_compare}
\end{figure}

Fig~\ref{fig:quality_traj_compare} compares the estimated trajectory and ground truth trajectory on the 3D colon print between DSO~\cite{DSO}, our framework using SuperPoint~\cite{super_point} and our proposed method. The pose estimation from the network is of arbitrary scale. Therefore, we first align the two trajectories using similarity transform~\cite{Kabsch:a12999} following with 
first-frame alignment for better visualization and comparison. Note that the estimated trajectories by our framework is more accurate with loops of similar shape as compared to the ground truth trajectory.

\subsection{Extra qualitative depth-map predictions results}\label{A. extra ct dm}
\begin{figure}[hptb]
    \centering
    \begin{tabular}[t]{ccc}

       \begin{tabular}{c}% if you add [t], than sub images are pushed down
             \parbox[t]{1mm}{\rotatebox[origin=c]{90}{Input}} \\\\\\\\\\\\\\\\\
             \parbox[t]{1mm}{\rotatebox[origin=c]{90}{DepthNet}}
        \end{tabular}
        \begin{subfigure}{.3\textwidth}
                \begin{tabular}{c}% if you add [t], than sub images are pushed down
                    \begin{subfigure}[t]{1\textwidth}
                        \centering
                       \includegraphics[width=1\textwidth]{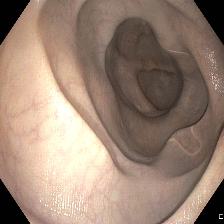}
                    \end{subfigure}\\
                    \begin{subfigure}[t]{1\textwidth}
                        \centering
                        \includegraphics[width=1\textwidth]{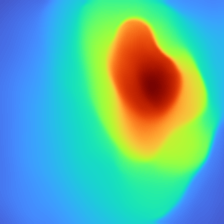}
                    \end{subfigure}\\
            \end{tabular}
        \end{subfigure}

        \begin{subfigure}{.3\textwidth}
                \begin{tabular}{c}% if you add [t], than sub images are pushed down
                    \begin{subfigure}[t]{1\textwidth}
                        \centering
                        \includegraphics[width=1\textwidth]{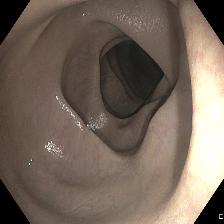}
                    \end{subfigure}\\
                    \begin{subfigure}[t]{1\textwidth}
                        \centering
                        \includegraphics[width=1\textwidth]{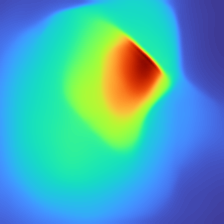}
                    \end{subfigure}\\
            \end{tabular}
        \end{subfigure}
        \begin{subfigure}{.3\textwidth}
                \begin{tabular}{c}% if you add [t], than sub images are pushed down
                    \begin{subfigure}[t]{1\textwidth}
                        \centering
                        \includegraphics[width=1\textwidth]{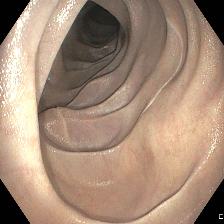}
                    \end{subfigure}\\
                    \begin{subfigure}[t]{1\textwidth}
                        \centering
                        \includegraphics[width=1\textwidth]{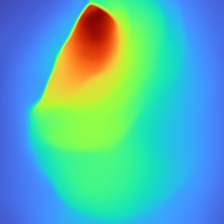}
                    \end{subfigure}\\
            \end{tabular}
        \end{subfigure}
    \end{tabular}
    \caption{Extra depth map prediction results on 3D colon print.}
\label{fig:extra_ndi_depth}
\end{figure}

\begin{figure}[hptb]
    \centering
    \begin{tabular}[t]{ccc}

       \begin{tabular}{c}% if you add [t], than sub images are pushed down
             \parbox[t]{1mm}{\rotatebox[origin=c]{90}{Input}} \\\\\\\\\\\\\\\\\
             \parbox[t]{1mm}{\rotatebox[origin=c]{90}{DepthNet}}
        \end{tabular}
        \begin{subfigure}{.3\textwidth}
                \begin{tabular}{c}% if you add [t], than sub images are pushed down
                    \begin{subfigure}[t]{1\textwidth}
                        \centering
                       \includegraphics[width=1\textwidth]{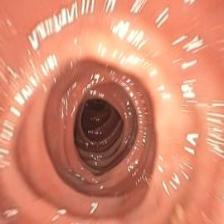}
                    \end{subfigure}\\
                    \begin{subfigure}[t]{1\textwidth}
                        \centering
                        \includegraphics[width=1\textwidth]{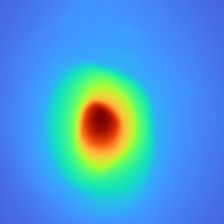}
                    \end{subfigure}\\
            \end{tabular}
        \end{subfigure}

        \begin{subfigure}{.3\textwidth}
                \begin{tabular}{c}% if you add [t], than sub images are pushed down
                    \begin{subfigure}[t]{1\textwidth}
                        \centering
                       \includegraphics[width=1\textwidth]{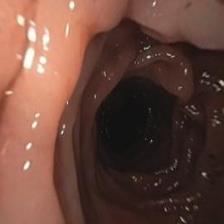}
                    \end{subfigure}\\
                    \begin{subfigure}[t]{1\textwidth}
                        \centering
                       \includegraphics[width=1\textwidth]{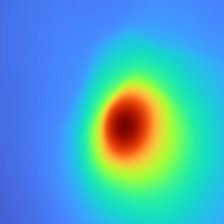}
                    \end{subfigure}\\
            \end{tabular}
        \end{subfigure}
        \begin{subfigure}{.3\textwidth}
                \begin{tabular}{c}% if you add [t], than sub images are pushed down
                    \begin{subfigure}[t]{1\textwidth}
                        \centering
                        \includegraphics[width=1\textwidth]{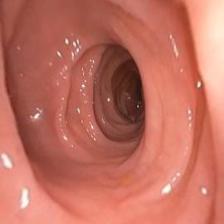}
                    \end{subfigure}\\
                    \begin{subfigure}[t]{1\textwidth}
                        \centering
                        \includegraphics[width=1\textwidth]{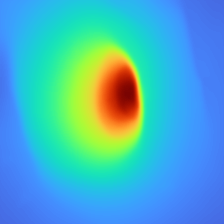}
                    \end{subfigure}\\
            \end{tabular}
        \end{subfigure}
    \end{tabular}
    \caption{Extra depth map prediction results on Colon10K~\cite{colon10k}. Left image exhibits a highly specular area with strong motion blur. The middle image exhibits strong illumination differences. The right image exhibits low texture images. In all three examples, our depth network produces detailed and artifact-free depth maps.}

\label{fig:extra_colon10k_depth}
\end{figure}

In Fig.~\ref{fig:extra_ndi_depth} and Fig.~\ref{fig:extra_colon10k_depth} we show extra depth-map predictions of the 3D colon print and Colon10K~\cite{colon10k}.

\subsection{Extra qualitative 3D \label{A. extra 3d qual} reconstruction results on Colon10K}
\begin{figure}[hptb]
    \centering
    \begin{tabular}[t]{cc}
    Colon10K~\cite{colon10k} & 3D colon print \\

        \begin{subfigure}{.45\textwidth}
                \begin{tabular}{c}% if you add [t], than sub images are pushed down
                    \begin{subfigure}[t]{1\textwidth}
                        \centering
                        \includegraphics[width=1\textwidth]{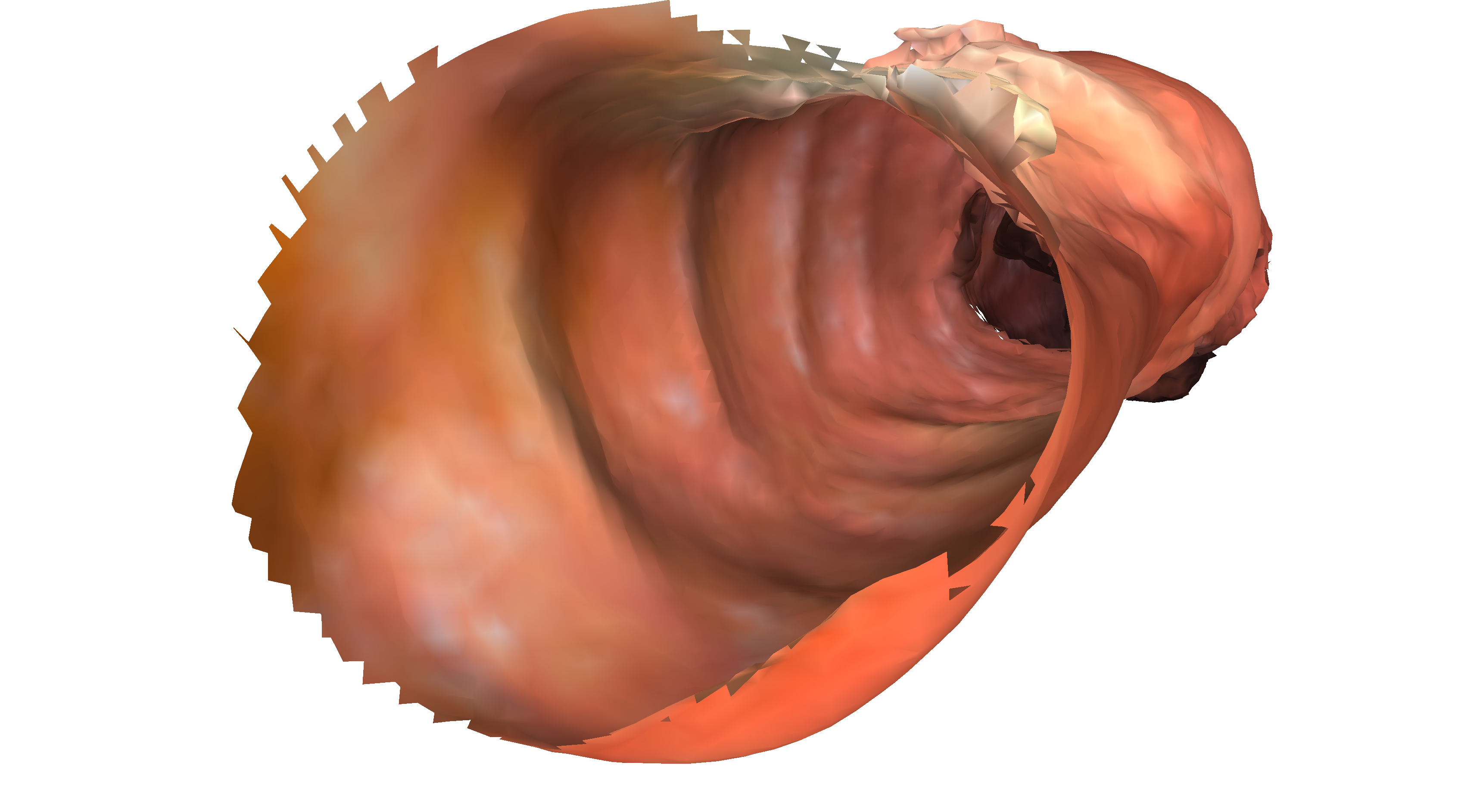}
                    \end{subfigure}\\
                    \begin{subfigure}[t]{1\textwidth}
                        \centering
                        \includegraphics[width=1\textwidth]{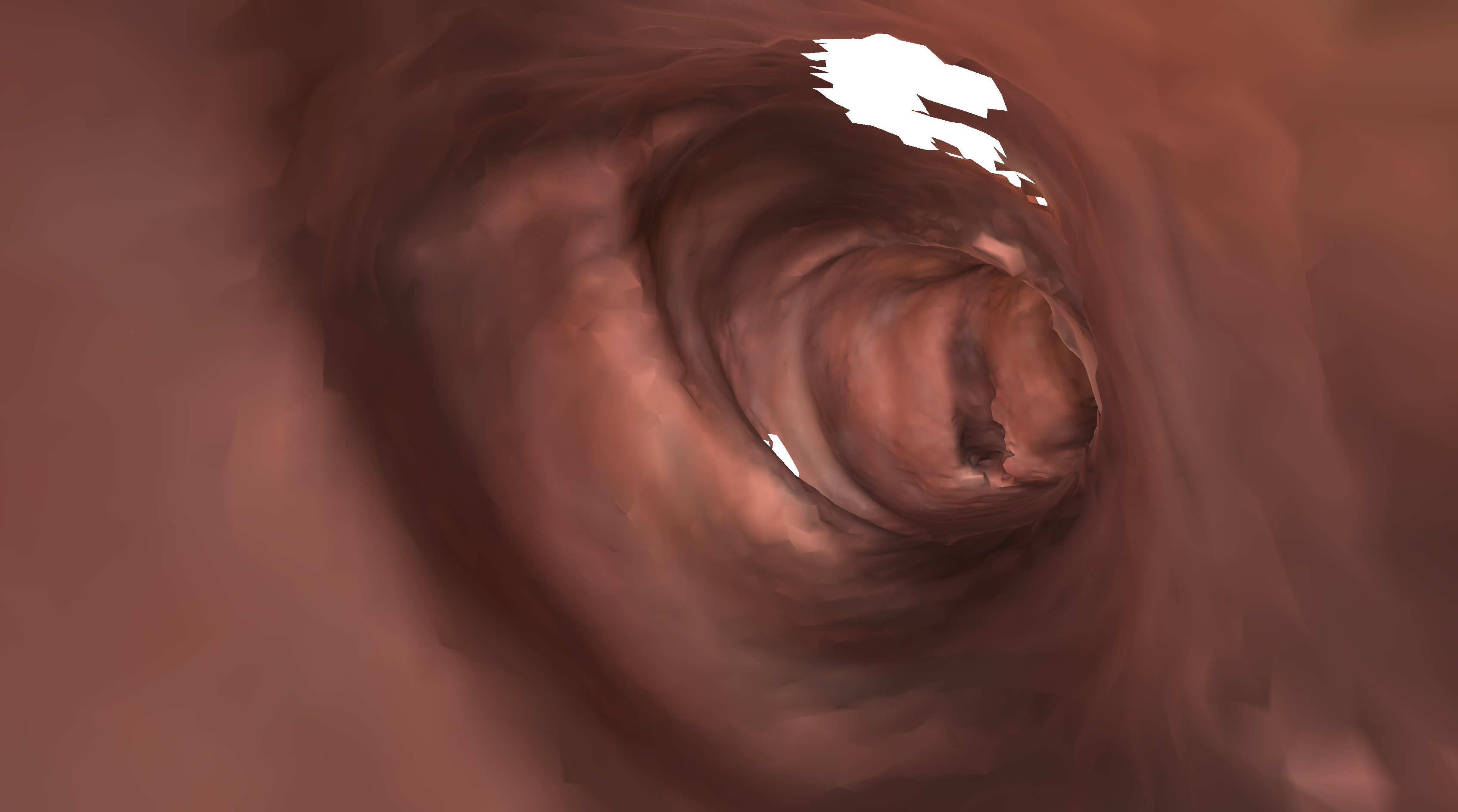}
                    \end{subfigure}\\
                    \begin{subfigure}[t]{1\textwidth}
                        \centering
                        \includegraphics[width=1\textwidth]{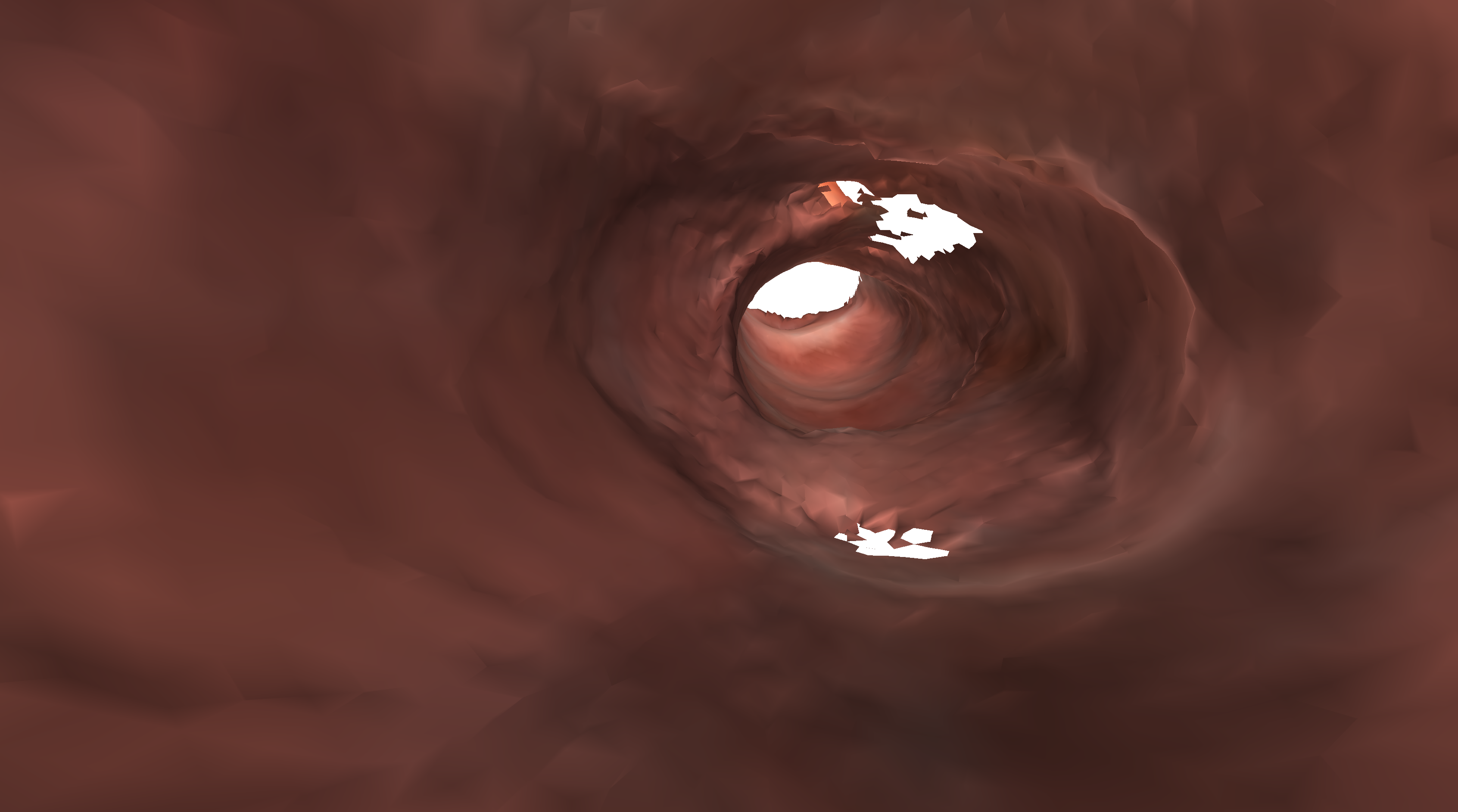}
                    \end{subfigure}\\

            \end{tabular}
        \end{subfigure} &
        \begin{subfigure}{.45\textwidth}
                \begin{tabular}{c}% if you add [t], than sub images are pushed down
                    \begin{subfigure}[t]{1\textwidth}
                        \centering
                        \includegraphics[width=1\textwidth]{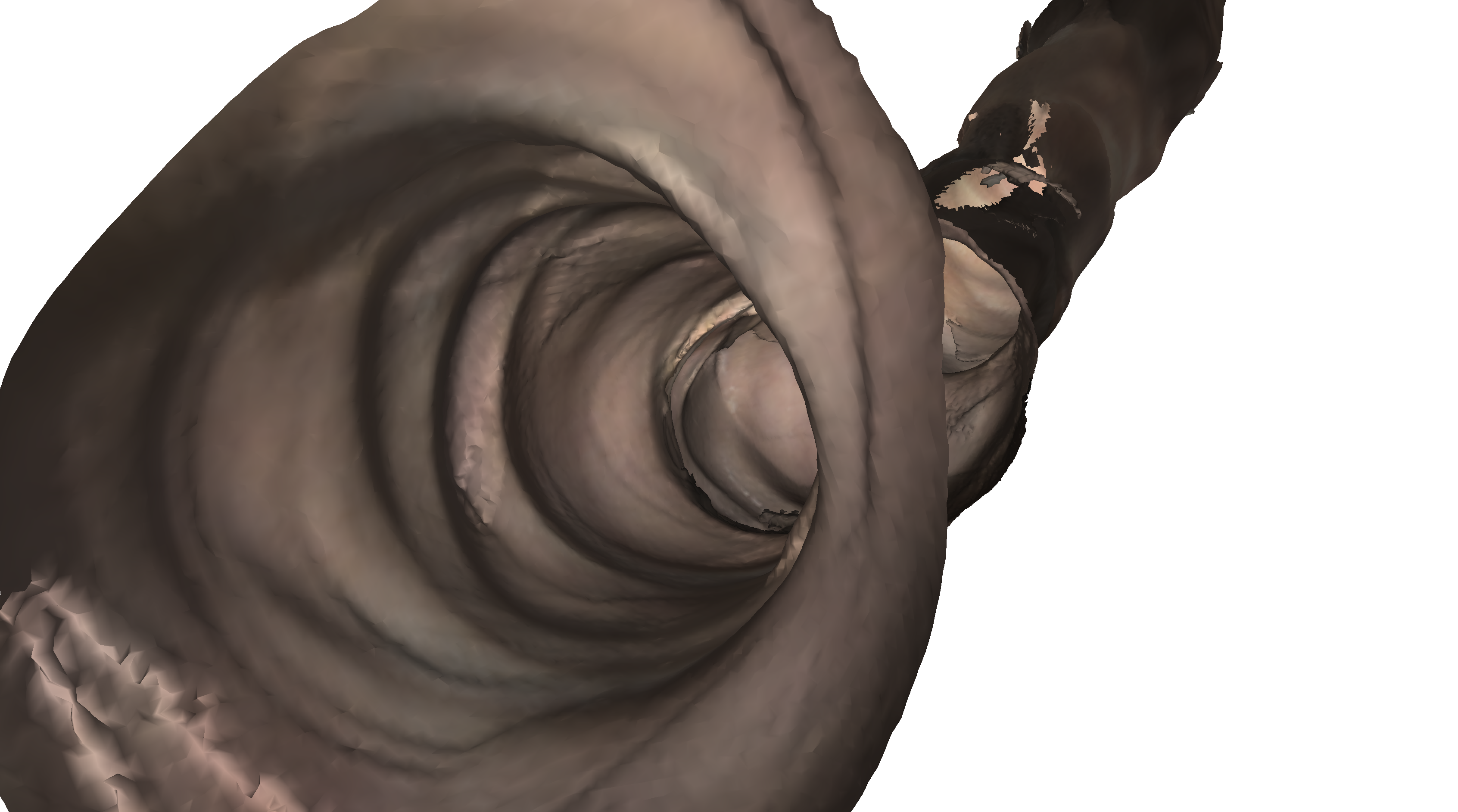}
                    \end{subfigure}\\
                    \begin{subfigure}[t]{1\textwidth}
                        \centering
                        \includegraphics[width=1\textwidth]{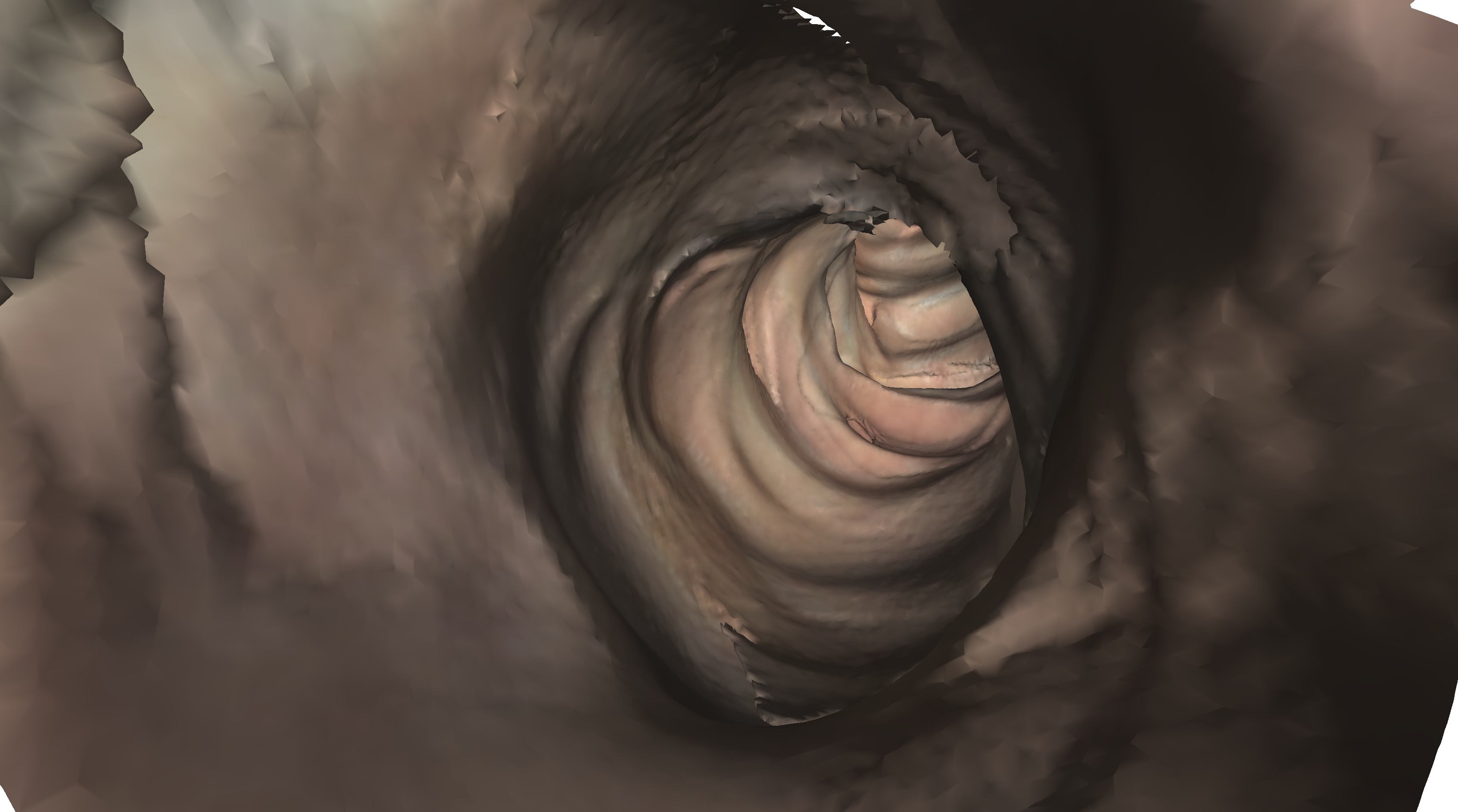}
                    \end{subfigure}\\
                    \begin{subfigure}[t]{1\textwidth}
                        \centering
                        \includegraphics[width=1\textwidth]{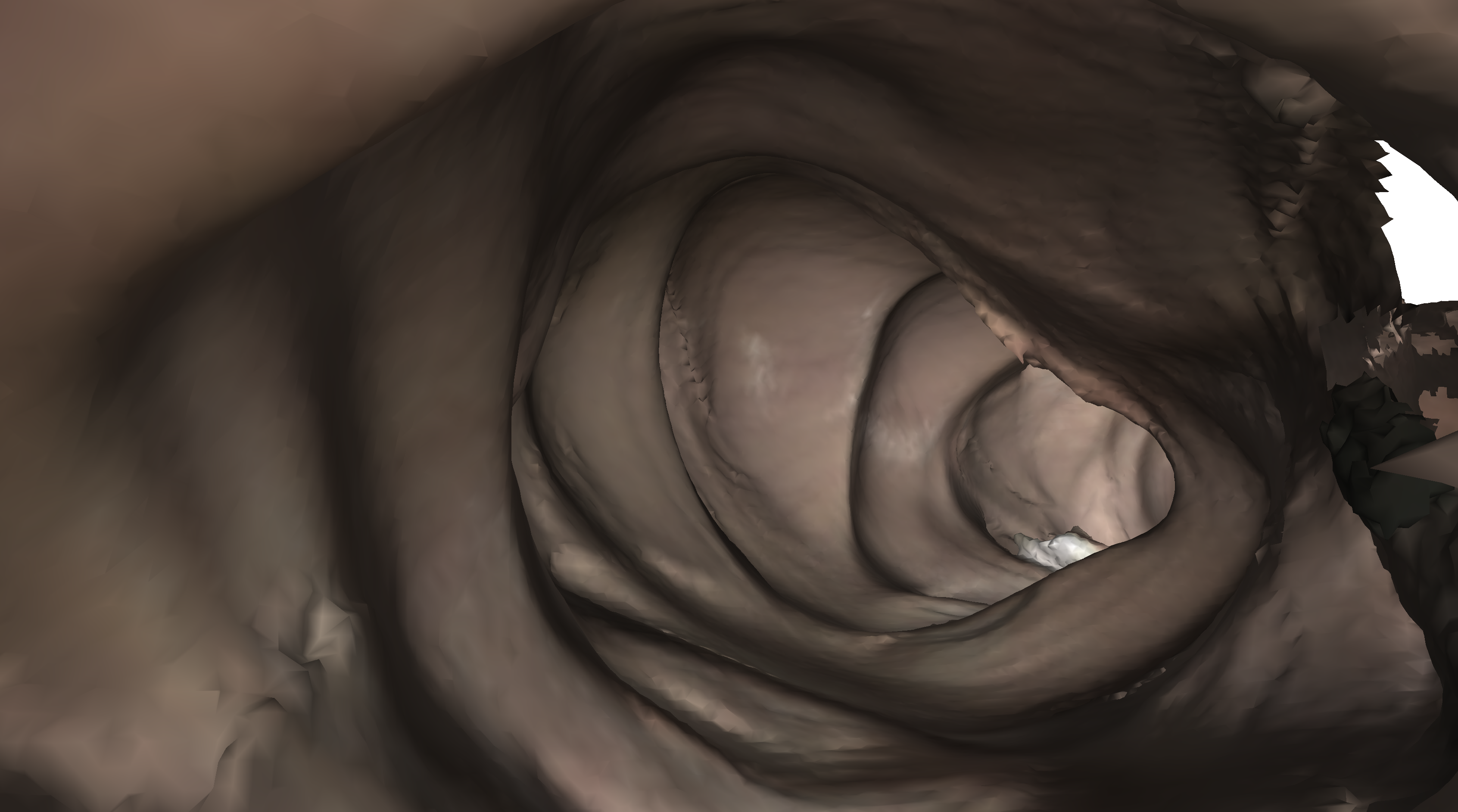}
                    \end{subfigure}\\

            \end{tabular}
        \end{subfigure} 
    \end{tabular}
    \caption{Extra reconstruction qualitative results on Colon10K~\cite{colon10k} and 3D colon print. }
\label{fig:extra_reco_quality}
\end{figure}

\begin{figure}[hptb]
    \begin{tabular}[t]{cc}

        \begin{subfigure}{0.4\textwidth}
                \begin{tabular}{c}% if you add [t], than sub images are pushed down
                % \smallskip
                
                    \begin{subfigure}[t]{1\textwidth}
                        \centering
                        \includegraphics[width=1\textwidth]{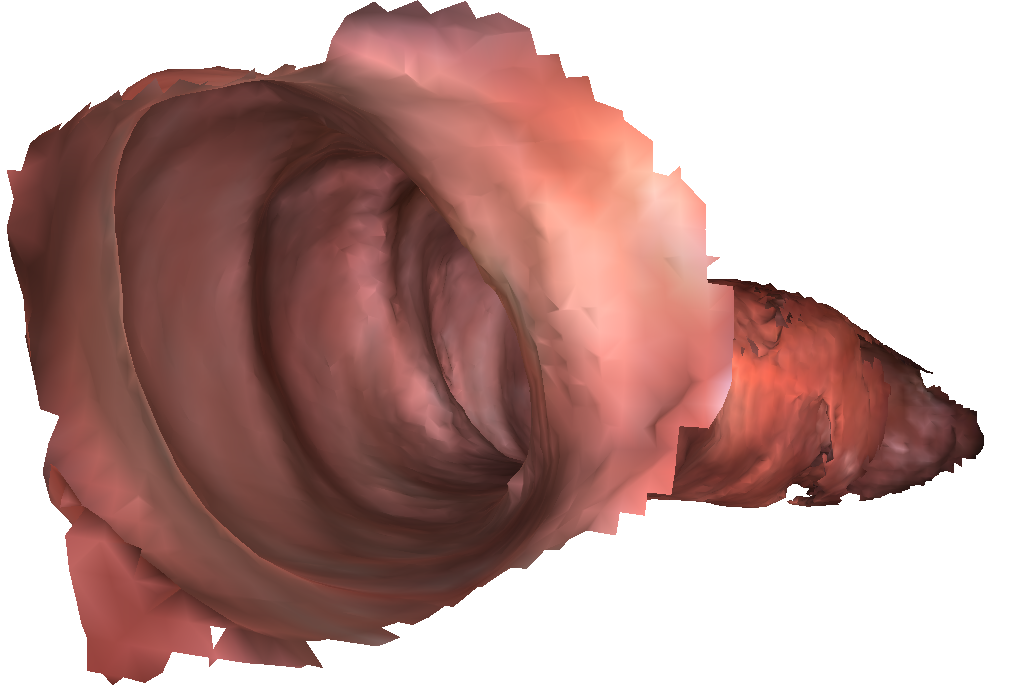}
                    \end{subfigure}\\
                    \begin{subfigure}[t]{1\textwidth}
                        \centering
                        \includegraphics[width=1\textwidth]{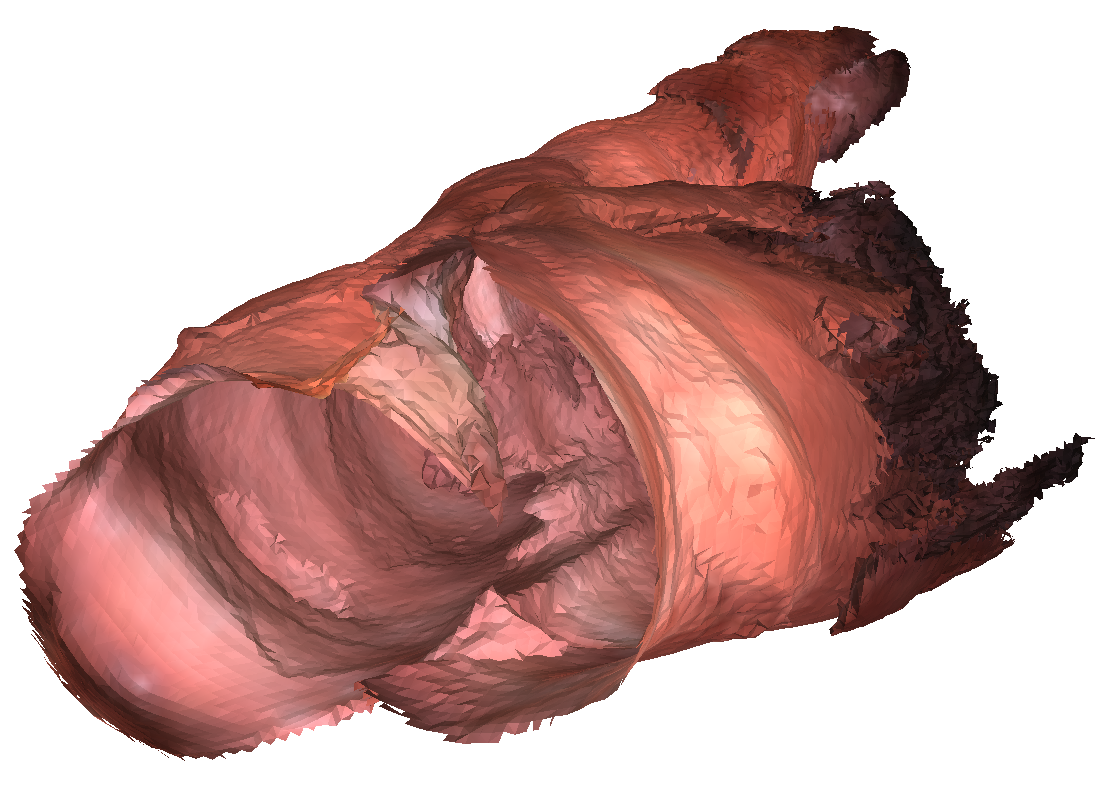}
                    \end{subfigure}\\

            \end{tabular}
            \caption{Seq.09}

        \end{subfigure}
        
        \begin{subfigure}{.45\textwidth}
                \begin{tabular}{c}% if you add [t], than sub images are pushed down
                \smallskip
                    \begin{subfigure}[t]{1\textwidth}
                        \centering 
                        \includegraphics[width=1\textwidth]{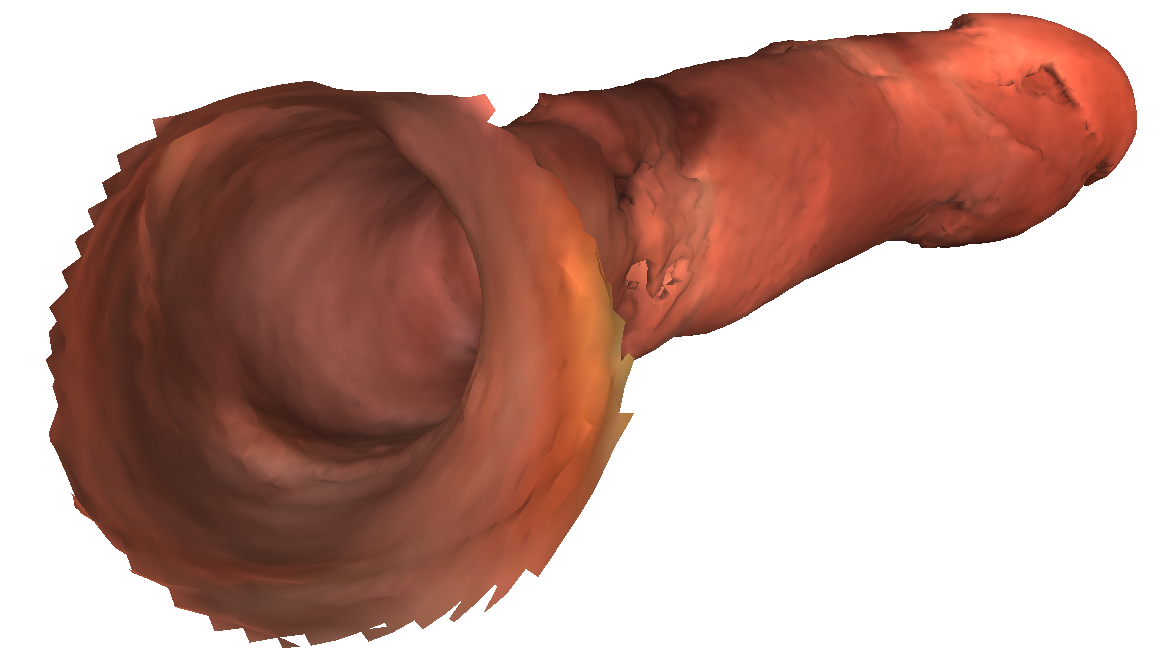}
                    \end{subfigure} \\
                    \begin{subfigure}[t]{1\textwidth}
                        \centering
                        \includegraphics[width=0.9\textwidth]{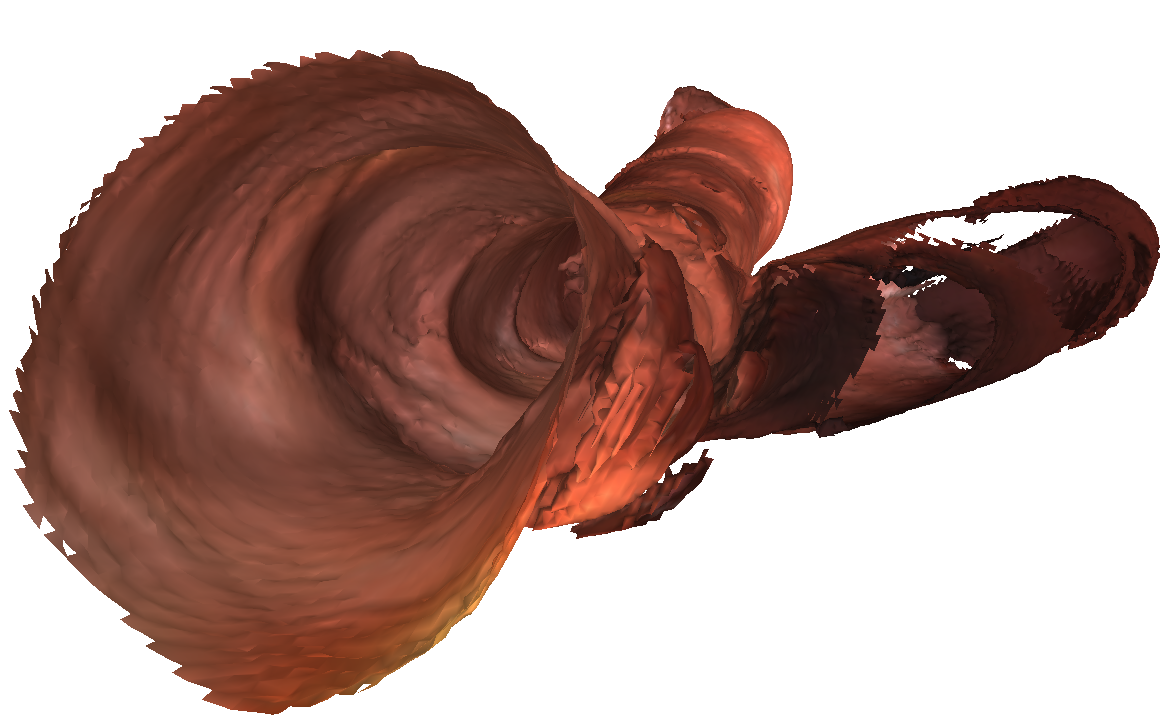}

                    \end{subfigure} \\

            \end{tabular}
            \caption{Seq.18}
            \vspace{-0.75cm}

        \end{subfigure}

    \end{tabular}
    \caption{Extra points-of-view of the 3D reconstruction results on Colon10K data-set. proposed framework (top), mesh reconstructed from depth and pose predictions by Godard et al.~\cite{monodepth2} (bottom)}
\label{fig:extra_reconstruction_qualitative_colon10k}
\end{figure}

In Fig.~\ref{fig:extra_reco_quality} and Fig.~\ref{fig:extra_reconstruction_qualitative_colon10k} we show extra points-of-view of the 3D reconstructions by our proposed framework on Colon10K~\cite{colon10k} and 3D colon print data.

\begin{table}[htbp]
\small
\begin{minipage}{1\textwidth}
\centering
        \caption{Synthetic data creation parameters
        }
        \label{tab:data_creation}
        \begin{tabular}{@{}ccccccccc@{}} \toprule
        \multicolumn{1}{l}{Parameters} & \multicolumn{8}{c}{Sequence} \\ 
        \cmidrule{2-9}
               & Seq. 1 & Seq. 2 & Seq. 3 & Seq. 4 & Seq. 5 & Seq. 6 & Seq. 7 & Seq. 8 \\ \midrule
          Path &
          A &
          RP &
          B &
          RP &
          RP &
          RP &
          RP &
          RP \\
        Trip Duration  & 150   & 150   & 150   & 150   & 150   & 150   & 150   & 150   \\
        Shots per sec  & 30    & 30    & 30    & 30    & 30    & 30    & 30    & 30    \\
        Shots Resolution &
          512x512 &
          512x512 &
          512x512 &
          512x512 &
          512x512 &
          512x512 &
          512x512 &
          512x512 \\
        Hue            & 0     & 0     & 3     & 3     & 100   & 100   & 100   & 100   \\
        Saturation     & 72    & 72    & 90    & 90    & 78    & 78    & 78    & 78    \\
        Value          & 100   & 100   & 100   & 100   & 52    & 52    & 52    & 52    \\
        Wetness        & 88    & 88    & 56    & 56    & 40    & 40    & 40    & 40    \\
        Vessel Size    & 60    & 60    & 60    & 60    & 60    & 60    & 60    & 60    \\
        Vessel Opacity & 30    & 30    & 30    & 30    & 30    & 30    & 30    & 30    \\
        Angle          & 150   & 150   & 150   & 150   & 150   & 150   & 150   & 150   \\
        Intensity      & 74    & 74    & 59    & 59    & 62    & 62    & 62    & 62    \\
        Distance       & 0     & 0     & 0     & 0     & 0     & 0     & 0     & 0     \\
        Shadow         & On    & On    & On    & On    & On    & On    & On    & On    \\
        Dynamic        & Off   & Off   & Off   & Off   & Off   & Off   & Off   & Off   \\
        Field of View  & 110   & 110   & 110   & 110   & 110   & 110   & 110   & 110   \\
        Vignette       & On    & On    & On    & On    & On    & On    & On    & On    \\
        Bloom          & On    & On    & On    & On    & On    & On    & On    & On    \\
        Grain          & 0     & 0     & 1     & 1     & 3     & 3     & 3     & 3    \\
        \bottomrule
\end{tabular}

\end{minipage}\hfill % maximize space between the minipages
\end{table}
\subsection{SuperPoint training}
SuperPoint~\cite{super_point} was trained using~\cite{superpoint_pytorch} Pytorch implementation with their suggested improvements that enable end to end training using a softargmax at the detector head and a sparse descriptor loss that allows an efficient training. Photo-metric augmentations were adapted to the colon data-set by lowering the contrast, blur and noise levels to values that enabled the extraction of features even from deeper shadowed areas of the colon.
the network was trained for about 100 epochs, with a batch size of 10 and learning rate of 0.0003. The best checkpoint was chosen based on validation set precision and recall.

\subsection{Supplementary Video Results}
\begin{figure}[hptb]
    \centering
    \begin{tabular}[t]{cc}
    RGB Image & Re-rendered reconstructed\\
              & model (with texture)\\
        \begin{subfigure}{.45\textwidth}
                \begin{tabular}{c}% if you add [t], than sub images are pushed down
                    \begin{subfigure}[t]{1\textwidth}
                        \centering
                        \includegraphics[width=1\textwidth]{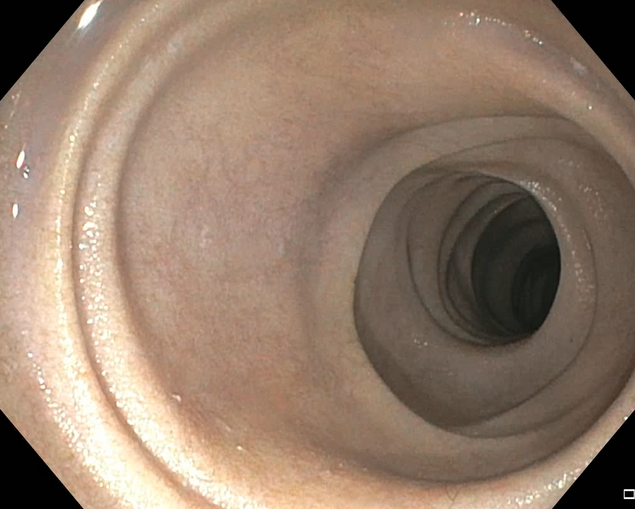}
                    \end{subfigure}\\
             \end{tabular}
        \end{subfigure} &
        \begin{subfigure}{.45\textwidth}
                \begin{tabular}{c}% if you add [t], than sub images are pushed down
                    \begin{subfigure}[t]{1\textwidth}
                        \centering
                        \includegraphics[width=1\textwidth]{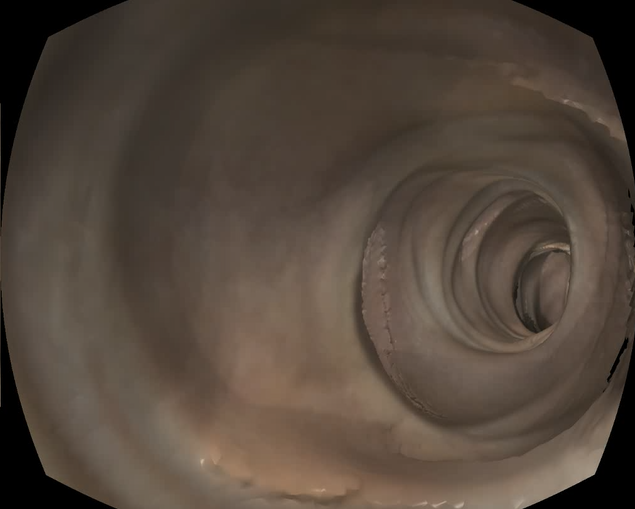}
                    \end{subfigure}\\
            \end{tabular}
            
        \end{subfigure}
    \end{tabular}
   \caption{Captured video and re-rendered reconstruction model similarity}
\label{fig:rgb_tex_geo}
\end{figure}

In the supplementary video, labeled as \textit{rgb\_tex\_geo.mp4}, we show the fully endoscopic investigation of the 3D colon print while comparing the resemblance between the reconstructed model and the captured RGB images. This is accomplished by re-rendering the reconstructed model using the camera intrinsics, camera predicted pose and framework's output mesh. In the video \textit{rgb\_tex\_geo.mp4} we visualise the captured video (Left) next to the re-rendered reconstructed model with texture (right). An example can be seen in Fig.~\ref{fig:rgb_tex_geo}. 
An additional camera fly-through video is available, labeled as \textit{fly\_through.mkv}, showing the final reconstruction of the 3D colon print.

\clearpage

\end{document}